\documentclass[letterpaper]{article} 
\usepackage[]{aaai2026}  
\usepackage{times}  
\usepackage{helvet}  
\usepackage{courier}  
\usepackage[hyphens]{url}  
\usepackage{graphicx} 
\usepackage{natbib}  
\usepackage{caption} 
\usepackage{subcaption} 
\usepackage{algorithm}
\usepackage{algorithmic}
\usepackage{amssymb}

\usepackage{booktabs}
\usepackage{multirow}
\usepackage[utf8]{inputenc} 
\usepackage[T1]{fontenc}    
\usepackage{url}            
\usepackage{booktabs}       
\usepackage{amsfonts}       
\usepackage{nicefrac}       
\usepackage{microtype}      
\usepackage{xcolor}         

\usepackage{microtype}
\usepackage{graphicx}
\usepackage{booktabs} 

\usepackage{subcaption}

\usepackage{amsmath}
\usepackage{amssymb}
\usepackage{mathtools}
\usepackage{amsthm}
\usepackage{bbding}

\usepackage{multirow}

\usepackage{newfloat}
\usepackage{listings}
\DeclareCaptionStyle{ruled}{labelfont=normalfont,labelsep=colon,strut=off} 
\floatstyle{ruled}
\newfloat{listing}{tb}{lst}{}
\floatname{listing}{Listing}
\title{Towards Federated Clustering: A Client-wise Private Graph Aggregation Framework}

\author{
    Guanxiong He,
    Jie Wang,
    Liaoyuan Tang,
    Zheng Wang\thanks{Corresponding author.},
    Rong Wang,
    Feiping Nie\\
}
\affiliations{
    School of Artificial Intelligence, Optics and Electronics (iOPEN), Northwestern Polytechnical University\\

	Xi'an Shaanxi, 710072, P.R.China\\
    heguanx@mail.nwpu.edu.cn,
    jiewang.dl@mail.nwpu.edu.cn,
    tangly@mail.nwpu.edu.cn,
    zhengwangml@gmail.com,
    wangrong07@tsinghua.org.cn,
    feipingnie@gmail.com
    
}

\usepackage{bibentry}

\begin{document}

\maketitle

\begin{abstract}
  Federated clustering addresses the critical challenge of extracting patterns from decentralized, unlabeled data. However, it is hampered by the flaw that current approaches are forced to accept a compromise between performance and privacy: \textit{transmitting embedding representations risks sensitive data leakage, while sharing only abstract cluster prototypes leads to diminished model accuracy}. To resolve this dilemma, we propose Structural Privacy-Preserving Federated Graph Clustering (SPP-FGC), a novel algorithm that innovatively leverages local structural graphs as the primary medium for privacy-preserving knowledge sharing, thus moving beyond the limitations of conventional techniques. Our framework operates on a clear client-server logic; on the client-side, each participant constructs a private structural graph that captures intrinsic data relationships, which the server then securely aggregates and aligns to form a comprehensive global graph from which a unified clustering structure is derived. The framework offers two distinct modes to suit different needs. SPP-FGC is designed as an efficient one-shot method that completes its task in a single communication round, ideal for rapid analysis. For more complex, unstructured data like images, SPP-FGC+ employs an iterative process where clients and the server collaboratively refine feature representations to achieve superior downstream performance. Extensive experiments demonstrate that our framework achieves state-of-the-art performance, improving clustering accuracy by up to 10\% (NMI) over federated baselines while maintaining provable privacy guarantees.
\end{abstract}

\begin{links}
    \link{Code}{https://github.com/GuanxiongHe23/FedGraph}
    \link{Extended version}{https://arxiv.org/}
\end{links}

\section{Introduction}
Federated Learning (FL) offers a promising approach to balancing data privacy with utilizing large-scale datasets for machine learning \cite{mcmahan2017communication, mcmahan2016federated, yang2019federated}. FL methods are generally divided into three transmission strategies: \textit{Model Averaging}, \textit{Embedding Sharing}, and \textit{Prototype Aggregation}, each offering different trade-offs among performance, communication cost, and privacy. Model Averaging preserves privacy effectively by aggregating parameters from locally trained models but incurs high communication overhead due to the complexity of model structures \cite{mcmahan2017communication, reddi2021adaptive}. Embedding Sharing reduces communication by transmitting intermediate representations but risks leaking sensitive information \cite{tan2022federated, wang2024turbosvm}. Prototype Aggregation further minimizes communication and privacy risks by sharing only representative prototypes, although this limits the ability to fully capture complex data structures \cite{tan2022fedproto, huang2023rethinking}. Thus, choosing the appropriate strategy is essential to effectively balancing privacy protection, communication efficiency, and performance.

Most FL research has emphasized supervised tasks, presuming access to labeled data, which is not always realistic \cite{wu2023faster, hu2024fedmut}. Clients may lack labels for their data, making supervised FL methods ineffective. This increased interest in unsupervised learning, particularly clustering, which groups data based on inherent similarities without needing labels \cite{dennis2021heterogeneity, xie2023fed}. Federated Clustering (FC) \cite{pan2022machine, nardi2024federated} addresses this need by enabling decentralized clustering while preserving data privacy, providing significant advantages over traditional centralized approaches \cite{basagni1999distributed, chen2016communication}. FC is especially valuable in privacy-sensitive sectors like healthcare and finance \cite{li2022secure}, with practical applications including Human Activity Recognition \cite{presotto2022fedclar} and Electricity Consumption Pattern Extraction \cite{wang2022federated}. By facilitating collaboration among isolated datasets, FC uncovers hidden patterns and relationships, becoming an important tool for unsupervised learning in industries where data sharing is restricted.

\begin{figure*}[t]
	\centering
	\includegraphics[width=0.75\textwidth]{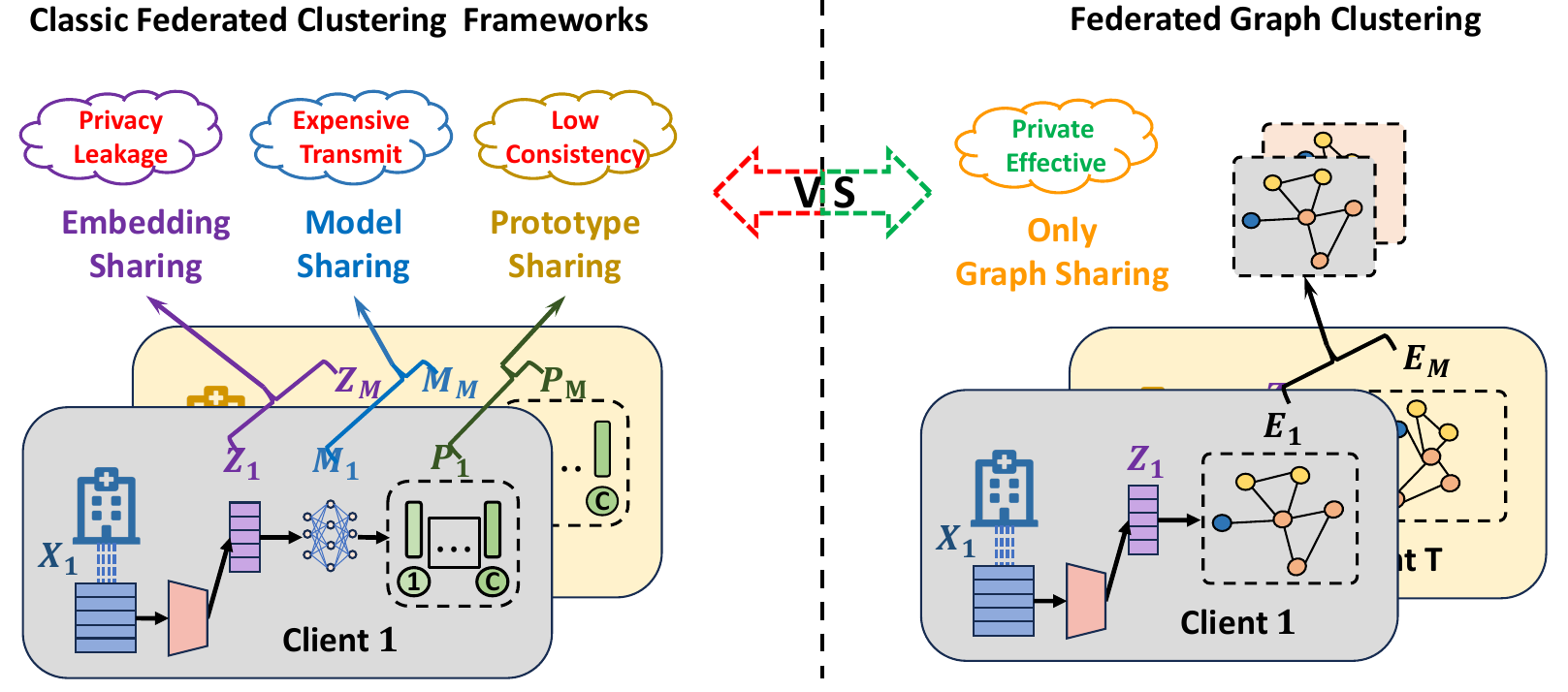}
	\caption{Classic FC paradigms and the proposed Federated Graph Clustering framework. The left shows the paradigms of Model Averaging, Embedding Sharing, and Prototype Aggregation, while the right highlights our graph-based approach.}
	\label{fig:1}
\end{figure*}

As shown in Fig \ref{fig:1}, current FC methods fall into three main paradigms. In Model Averaging, clients share locally trained model weights, achieving strong performance but incurring high communication costs \cite{mashhadi2021deep, li2023differentially}. The Embedding Sharing paradigm has clients send feature embeddings to the server, which allows knowledge sharing but risks leaking private information \cite{chen2023federated, qiao2024federated}. In Prototype Aggregation, clients share only local cluster centers, a method that reduces communication and improves privacy but can compromise clustering quality and consistency \cite{dennis2021heterogeneity, pedrycz2021federated}. Each paradigm offers a different trade-off between performance, privacy, and complexity. We aim to strike a better balance among these factors while addressing the limitations of existing methods.

To achieve this balance, we propose a new method, SPP-FGC. Clients build structural graphs from their local data and send them, rather than raw data or embeddings, to the server, preserving privacy. During communication, only similarity graphs and prototype distributions are shared, and these are further protected using Differential Privacy (DP) techniques \cite{dwork2006differential, zhao2022survey} to reduce leakage risk. The server aggregates these similarity graphs and uses the Constrained Laplacian Rank method \cite{nie2016constrained} to refine the global structure. It then identifies clusters and sends global assignments back to clients. To better handle complex data, we extend the method to SPP-FGC+, which includes feature learning and iterative refinement. In this version, clients use Variational Autoencoders (VAE) \cite{doersch2016tutorial, xie2016unsupervised} with Deep Embedded Clustering (DEC) loss for self-supervised feature extraction. As the global structure updates through graph aggregation, new cluster prototypes are sent to guide further learning. This iterative process improves feature representation and clustering accuracy over time, making the method more adaptable to complex and dynamic datasets. The key contributions of our proposed methods are highlighted from the following three distinct perspectives.

\begin{itemize}
	\item  We introduce a novel method that leverages local structural graphs as the knowledge-sharing medium, effectively overcoming the privacy risks of sharing embeddings and the performance loss of sharing only prototypes.
	
	\item  Our framework ensures security by abstracting client data into local private structures and employing differential privacy to facilitate their secure aggregation on the server.

	\item Our method is highly adaptable to diverse scenarios, offering an efficient one-shot mode (SPP-FGC) for simple datasets and an iterative version (SPP-FGC+) that achieves superior performance on complex, unstructured data through collaborative learning.
\end{itemize}

\section{Related Works}

\paragraph{Federated Clustering:} FC follows three main FL paradigms. Under Model Averaging, clients train local clustering models and send their parameters to a central server to build a global model \cite{mcmahan2017communication}. This approach protects privacy and achieves strong performance but incurs high communication costs. For example, F-DEC \cite{mashhadi2021deep} integrates deep embedding clustering into Fed-Avg by sharing weighted local models and updating via gradients; FednadamN \cite{hasan2025autoencoder} adds Adam and Nadam optimizers to speed convergence; and PPFC-GAN \cite{yan2023privacy} uses GANs \cite{2020Goodfellowgan} to generate private data samples for global clustering. In Embedding Sharing, clients send feature embeddings instead of full models—FedDMVC \cite{chen2023federated} employs contrastive learning for robust embeddings, and FedCA \cite{zhang2023federated} uses contrastive averaging with a dictionary and alignment strategy. Finally, Prototype Aggregation has clients share cluster centers as prototypes: k-FED \cite{dennis2021heterogeneity} extends k-means by aggregating centroids from each client, and FFCM \cite{pedrycz2021federated} adds a fuzzy extension for greater robustness. Different from the above methods, the graph-based framework of SPP-FGC  exchanges only private graphs, enhanced with differential privacy, to capture deeper relationships and deliver more consistent global clusters without excessive overhead.

\paragraph{Differential Privacy:} As a sophisticated mathematical framework designed to protect individual privacy within large datasets while still allowing meaningful aggregate analysis, DP often takes Gaussian and Laplace mechanisms into practice \cite{dwork2006differential, abadi2016deep}. It is widely used in FL to protect individual users' data by adding noise to model updates, ensuring that sensitive information cannot be inferred from the aggregated results \cite{wei2020federated, hu2020personalized, triastcyn2019federated}.

\section{Problem Formulation}

FL is a decentralized machine learning mechanism that allows multiple clients to collaboratively train a global model without sharing their local data. It addresses privacy and data ownership concerns by keeping data on the client side and only exchanging non-sensitive information. In the scenario of $M$ clients, each one has their own dataset $D_1,...,D_M$ where $D_i=\{(v_j,y_j)\}_{j=1}^{N_i}$, with $v\in\mathbb{R}^d$ representing $d$-dimensional features and $y\in\{1,2...,C\}$ representing labels across $C$ classes. The loss function $\mathcal{\ell}(v, y, \theta)$ measures the loss for a sample $(v,y)$ with model parameters $\theta$. The combined dataset of all clients is $\textbf{D} = \bigcup_{i=1}^MD_i$ and the goal of FL is to learn a model $\theta^*$ that minimizes:
\begin{equation}
	\underset{\theta}{\arg\min}\sum_{i=1}^{M}\frac{|D_i|}{|\textbf{D}|}\mathbb{E}_{(v,y)\sim D_i}\mathcal{\ell}(v,y,\theta).
	\label{eq:1}
\end{equation}

FC follows a similar objective on unlabeled data $\bar{D}_i=\{v_j\}_{j=1}^{N_i}$. It aims to optimize data clustering within each client by utilizing information from all clients, thereby enhancing clustering quality and consistency without compromising privacy. The optimization goal in FC is to minimize the clustering loss $\mathcal{\ell}_c$ across all clients:
\begin{equation}
	\underset{\theta_1,...,\theta_M}{\arg\min}\sum_{i=1}^{M}\frac{|\bar{D}_i|}{|\bar{\textbf{D}}|}\mathbb{E}_{v\sim \hat{D}_i}\mathcal{\ell}_c(v,\theta).
	\label{eq:2}
\end{equation}

The main challenge in FC is achieving high-quality, collaborative clustering while maintaining strict privacy standards. Current methods often have to sacrifice either performance or privacy to address this balance. Therefore, we aim to find an optimal balance by transmitting similarity graphs instead of raw data, ensuring both effective clustering and robust privacy protection.

\section{Methodology}

To address the challenge of achieving high-quality federated clustering while ensuring strong privacy protection, we introduce Structural Privacy-Preserving Federated Graph Clustering (SPP-FGC). This graph-based approach enhances data privacy by utilizing similarity graphs instead of raw data features, along with the DP mechanism. Additionally, we extend this framework with the SPP-FGC+ variant, which integrates deep feature extraction and iterative optimization. This enhancement enables adaptive and powerful feature discovery, making SPP-FGC+ suitable for complex, real-world data.

\begin{figure*}[!t]
	\centering
	\includegraphics[width=1\textwidth]{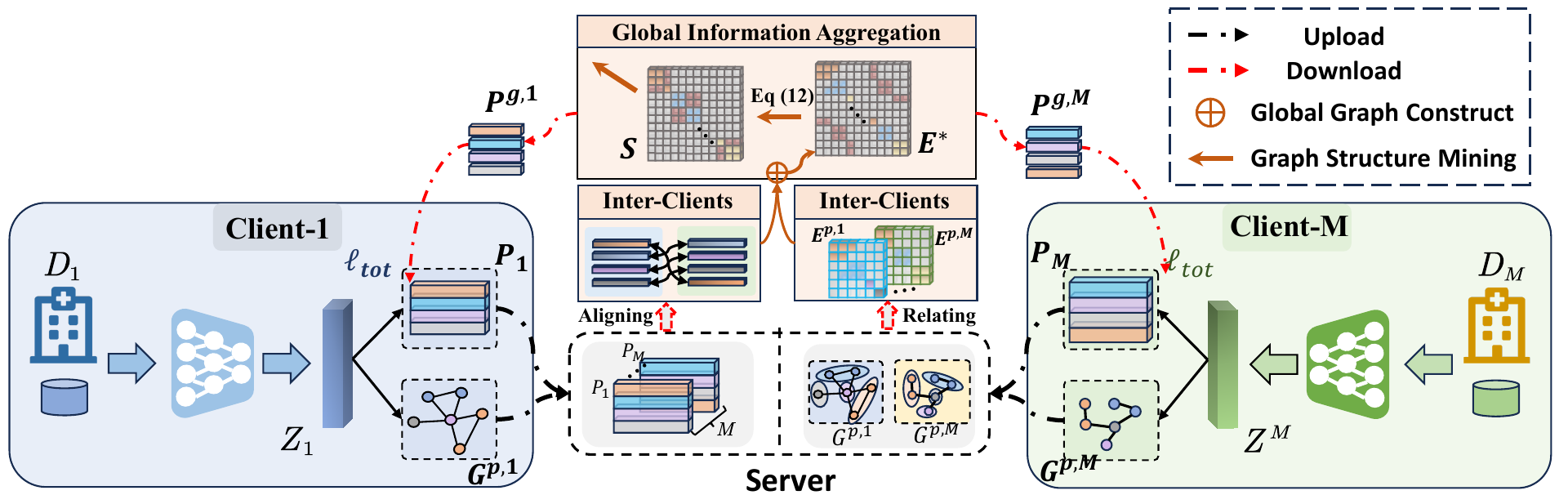}
	\caption{Overview of SPP-FGC+. Clients generate private structural graphs from learned features and upload them to the server. The server fuses these inputs into a global graph, derives new cluster prototypes, and sends them back as feedback.}
	\label{fig:2}
\end{figure*}

In our SPP-FGC method, we draw from traditional graph clustering, which typically involves two steps: building a similarity graph from the data and using it for clustering. This approach fits well within the FL framework, where data stays decentralized on clients, and only the similarity graphs are shared with the central server. The goal of SPP-FGC is to combine the clients' ability to capture local data relationships with the server's role in aggregating this information for better global clustering. By integrating data from all clients, SPP-FGC strengthens cluster connections and corrects local errors. Our method tackles three key challenges: \textbf{accurately mining} structure on each client, \textbf{securely transmitting} information to the server, and \textbf{effectively merging} data from all clients to enhance overall clustering.

\subsection{Private Structural Graph Construction:} To uncover the underlying structure in each client's data, we construct a private graph $G^p=(V^p, E^p)$ that captures the neighborhood relationships of the samples. In this graph, similar samples are placed closer together, and dissimilar ones farther apart. This approach enhances the representation of data relationships, leading to better clustering results. To create a more informative graph, we define the following optimization objective:
\begin{equation}
	\begin{aligned}
		&\underset{E^p}{\min} \sum_{i,j=1}^{N} \left( ||v_i^p - v_j^p||_2^2 E_{ij}^p + \gamma (E_{ij}^p)^2 \right). \\
		\text{s.t. } \forall i, \ &(E_{i}^p)^{\text{T}} \mathbf{1} = 1, \ 0 \leq E_{ij}^p \leq 1, \ \text{rank}(L_{E^p}) = N - C
	\end{aligned}
	\label{eq:3}
\end{equation}
Here, $E^p$ represents the relationship matrix for specified clients, each element $E_{ij}^p$ measures the closeness of the relationship between the $i$-th and $j$-th private samples vectors $v_i^p,v_j^p$. Meanwhile, $E_i^p$ represents the relationships between a given sample and all others. Additionally, $L_{E^p}$ is the Laplacian matrix of $E^p$, defined as:
\begin{equation}
	L_{E^p} = D_{E^p} - \frac{(E^p)^{\text{T}} + E^p}{2},
	\label{eq:4}
\end{equation}
where $D_{E^p}$ denotes the degree matrix of $E^p$. Here are two constraints: 1) The rank constraint on $L_{E^P}$ forces the graph to have exactly $C$ connected components. The objective can be understood as obtaining the optimal private structural matrix $E^P$ by minimizing the combined effect of the Euclidean distance between feature vectors. 2) The L2 regularization term $\gamma (E_{ij}^p)^2$ is taken to avoid the trivial solution where each data point connects only to its single nearest neighbor. It encourages a smoother distribution of connection weights across multiple neighbors.

To make the rank constraint in Eq.\ref{eq:3} more tractable, we reformulate the problem by leveraging Ky Fan’s Theorem \cite{fan1949theorem}. This involves introducing a continuous cluster indicator matrix $F \in \mathbb{R}^{N \times C}$. This matrix serves as a relaxed representation of cluster assignments, where its columns are the eigenvectors corresponding to the C smallest eigenvalues of $L_{E^P}$. The problem can be reformulated as:
\begin{equation}
	\begin{aligned}
		\underset{E^p, F}{\min} \sum_{i,j=1}^{N} \left( ||v_i^p - v_j^p||_2^2 E_{ij}^p + \gamma (E_{ij}^p)^2 \right) &+ 2\lambda \text{Tr}(F^{\text{T}} L_{E^p} F). \\
		\text{s.t. } \forall i, \ (E_{i}^p)^{\text{T}} \mathbf{1} = 1, \ 0 \leq E_{ij}^p &\leq 1, \ F^{\text{T}} F = I
	\end{aligned}
	\label{eq:5}
\end{equation}
To solve the simplified problem, the parameters $F$ and $E^p$ can be optimized alternately. First, fix $E^p$, the optimal solution of the indicating matrix $F$ is formed by the $C$ eigenvectors of $L_{E^p}$ corresponding to the $C$ smallest eigenvalues.

Then, with $F$ fixed, using the Lagrangian function and Karush-Kuhn-Tucker (KKT) conditions, the optimal solution for $E_i^p$ is given by:
\begin{equation}
	E_{ij}^p = -\frac{d_{ij}}{2\gamma_i} + \eta,
	\label{eq:6}
\end{equation}
where $d_{ij} = d_{ij}^x + d_{ij}^f$, with $d_{ij}^x = ||v_i^p - v_j^p||_2^2$ and $d_{ij}^f = ||f_i - f_j||_2^2$. Here, $f_i$ denotes the row vector of $F$, also the indicating vector. The detailed process of the private structural graph optimization is provided in Appendix A.

For a specified client $k$, the private structural matrix $E^{p,k}$ is constructed through the above optimization. To capture the underlying data distribution, especially with heterogeneous data across clients, prototypes are generated for each cluster using a Gaussian Mixture Model (GMM). Specifically, for each cluster $c$ within client $k$, its prototype is defined as $\mathcal{N}(\mu_{k,c}, \Sigma_{k,c})$, where $\mu_{k,c}$ is the mean centroid of cluster $c$, and $\Sigma_{k,c}$  is the covariance matrix that captures the variance and correlations within the cluster. These prototypes summarize the statistical properties of each cluster, aiding in more effective aggregation at the central server.

\subsection{Private Information Transmission:} To ensure data privacy and secure the transmission of sensitive information in the SPP-FGC framework, we use DP techniques, specifically the Laplace mechanism, to add calibrated noise to the transmitted data. This approach ensures that the exchange of the prototype set $P_k=\{(\mu_c,\Sigma_c)\}_{c=1}^{C}$ adheres to $\epsilon$-DP, providing strong privacy protection against potential data leaks.

For prototypes on client $k$, the sensitivities $\Delta^k$ are:
\begin{equation}
	\begin{aligned}
		\Delta_\mu^k &= \max_{k,c} \| \mu_{kc} - \mu_{kc}^{'} \|_1, \\
		\Delta_\Sigma^k &= \max_{k,c} \| \Sigma_{kc} - \Sigma_{kc}^{'} \|_1,
	\end{aligned}
	\label{eq:7}
\end{equation}
where $\mu_{kc}^{'}$ and $\Sigma_{kc}^{'}$ are the mean and covariance matrices of the modified dataset $D'$. Laplace noise is then added to the parameters to ensure differential privacy:
\begin{equation}
	\begin{aligned}
		\bar{\mu}_{i,c} &= \mu_{i,c} + \text{Lap}\left(\frac{\Delta_\mu}{\epsilon}\right), \\
		\bar{\Sigma}_{i,c} &= \Sigma_{i,c} + \text{Lap}\left(\frac{\Delta_\Sigma}{\epsilon}\right). \\
	\end{aligned}
	\label{eq:9}
\end{equation}

Here, $\text{Lap}(\cdot)$ represents the Laplace distribution with scale parameter $b = \Delta/\epsilon$, where $\Delta$ is the sensitivity and $\epsilon$ is the privacy budget. Finally, the $k$-th private prototypes $\bar{P}_k=\{\bar{\mu}_{k,c}, \bar{\Sigma}_{k,c}\}_{c=1}^C$ are securely transmitted to the central server. Details about privacy sensitivity can be seen in Appendix C.

\subsection{Global Information Aggregation:} Upon receiving the Laplace-encrypted local prototypes $\bar{P}_k$ and private structural matrix $E^{p,k}$ from all $M$ clients, the central server faces the crucial task of mining consistent clustering structures from the aggregated information. This step is essential to ensure that the global clustering reflects the collective insights derived from all clients' data while preserving data privacy. 

To achieve this, the central server constructs a global structural matrix $E^*$, which encapsulates both the local structural information and the inter-client similarities. The construction process begins by organizing $E^*$ as a block matrix, where each diagonal block corresponds to a client's private structural matrix $E^{p,k}$, thus retaining the localized clustering information of each client. Formally, $E^*$ is represented as:

\begin{equation}
	E^* = 
	\begin{bmatrix}
		E^{p,1} & E_{2}^{g,1} & \cdots & E_{M}^{g,1} \\
		E_{1}^{g,2} & E^{p,2} & \cdots & E_{M}^{g,2} \\
		\vdots & \vdots & \ddots & \vdots \\
		E_{1}^{g,M} & E_{2}^{g,M} & \cdots & E^{p,M} \\
	\end{bmatrix}.
	\label{eq:10}
\end{equation}

The off-diagonal blocks, $E_{j}^{g,i}$, represent the inter-client relation between clients $i$ and $j$. These relationships are computed by measuring the affinity between their respective cluster prototypes. Specifically, the similarity between any sample from client i and any sample from client j is determined by the Kullback-Leibler (KL) divergence between their assigned prototypes. Let $\bar{P}_i^{c(m)}$ be the prototype for the cluster assigned to sample $m$ in client $i$, and $\bar{P}_j^{c(n)}$ be the prototype for the cluster assigned to sample $n$ in client $j$. The KL divergence $\text{KL}_{i,j}(\bar{P}_i^{c(m)},\bar{P}_j^{c(n)})$ quantifies the dissimilarity between their underlying distributions.

To convert this divergence into a similarity score, we apply an exponential kernel, ensuring that greater divergence results in a lower similarity value. The inter-client relation graph $E_{j}^{g,i}$ is thus an $N_i \times N_j$ matrix where each element is computed as:

\begin{equation}
	E_{j}^{g,i} = \left[ \exp\left( - \text{KL}_{i,j}(\bar{P}_i,\bar{P}_j) \right) \right].
	\label{eq:11}
\end{equation}
This process effectively captures the pairwise alignment of cluster distributions across the federation. By assembling the private structural matrices $E^{p,k}$ and the inter-client relation graphs $E_{j}^{g,i}$ into the global matrix $E^*$, the server fuses localized knowledge with global relational insights, creating a robust foundation for identifying consistent, overarching cluster structures.

\begin{algorithm}[!t]
    \caption{SPP-FGC}
    \label{alg:SPP-FGC}
    \begin{algorithmic}[1]
        \STATE \textbf{Input:} $M$ Clients, $C$ Clusters, Privacy budget $\epsilon$
        \STATE
        \STATE \textbf{Client-side Procedures:}
        \FOR{each client $k = 1$ to $M$}
            \STATE Construct $E^{p,k} \gets \text{Eq.~\ref{eq:5}}$
            \STATE Compute $P_k \gets \text{GMM}(D_k)$
            \STATE Apply $\bar{P}_k \gets \text{Lap}(\epsilon, P_k)$
            \STATE Upload $E^{p,k}$ and $\bar{P}_k$ to Server 
        \ENDFOR
        \STATE
        \STATE \textbf{Server-side Procedures:}
        \STATE Compute $\{E_{j}^{g,i}\}_{i,j=1}^M \gets \text{Eq.~\ref{eq:11}}$
        \STATE Construct $E^* \gets \text{Aggregate}(E^{g}, E^{p})$
        \STATE Global aggregation $S \gets \text{Eq.~\ref{eq:13}}$
        \STATE Cluster assignments $\mathbf{A} \gets \text{K-Means}(S)$
        \STATE
        \STATE \textbf{Output:} Global cluster assignments $\mathbf{A}$
    \end{algorithmic}
\end{algorithm}

To further learn the global matrix $E^*$ and extract a decision similarity matrix $S$ with clear clustering structures, we impose a Laplacian rank restriction to enhance the structural features in $S$. The desired matrix has a block diagonal structure, where each block corresponds to a cluster. This is achieved by solving the following optimization problem:

\begin{equation}
	\begin{aligned}
		&\underset{S}{\min} \ ||S - E^*||_F^2. \\
		\text{s.t.} \quad& \text{rank}(L_S) = N - C
	\end{aligned}
	\label{eq:12}
\end{equation}
We apply the Lagrangian function and Ky Fan’s Theorem to transform problem into the following optimization problem:
\begin{equation}
	\begin{aligned}
		&\underset{S}{\min} \ ||S - E^*||_F^2 + 2\lambda \text{Tr}(F^{\text{T}} L_S F). \\
		\text{s.t.} \quad \sum_j &S_{ij} = 1, \ S_{ij} \geq 0, \ F \in \mathbb{R}^{N \times C}, \ F^{\text{T}} F = I
	\end{aligned}
	\label{eq:13}
\end{equation}

To solve this problem, we use the alternating optimization approach. Specifically, we iteratively optimize $F$ and $S$ by fixing one and solving for the other.

When $S$ is fixed, the optimal solution to the optimization problem is obtained by selecting the $C$ eigenvectors of $L_S$ corresponding to the $C$ smallest eigenvalues.

Using the KKT conditions, the optimal solution for $S$, when $F$ is fixed, is given by:
\begin{equation}
	S_{ij} = -\frac{d_{ij}}{2\gamma} + \eta.
	\label{eq:14}
\end{equation}

In summary, the primary framework of SPP-FGC is outlined in Algorithm \ref{alg:SPP-FGC}. The detailed proof can be seen in Appendix B.

\subsection{SPP-FGC with Self-Supervised Feature Learning}

\begin{algorithm}[!t]
    \caption{SPP-FGC+}
    \label{alg:SPP-FGC+}
    \begin{algorithmic}[1]
        \STATE \textbf{Input:} $T$ rounds, $M$ clients, $C$ clusters, privacy budget $\epsilon$
        \STATE Initialize VAE-DEC, obtain global prototypes $P_g^0$
        \STATE
        \FOR{$t = 1$ to $T$}
            \STATE \textbf{Client-side Procedures:}
            \FOR{each client $k = 1$ to $M$}
                \STATE Train VAE-DEC with $P_g^{t-1}$
                \STATE Construct $E^{p,k} \gets \text{Eq.~\ref{eq:5}}$ on $Z_k$
                \STATE Extract $P_k \gets DEC$
                \STATE Apply $\bar{P}_k \gets \text{Lap}(\epsilon, P_k)$
                \STATE Upload $E^{p,k}$ and $\bar{P}_k$ to server
            \ENDFOR
            \STATE
            \STATE \textbf{Server-side Procedures:}
            \STATE Compute $\{E_{j}^{g,i}\}_{i,j=1}^M \gets \text{Eq.~\ref{eq:11}}$
            \STATE Construct $E^* \gets \text{Aggregate}(E^g, E^{p})$
            \STATE Global aggregation $S \gets \text{Eq.~\ref{eq:13}}$
            \STATE Update $P_g^t \gets \text{K-Means}(S)$
        \ENDFOR
        \STATE
        \STATE \textbf{Output:} Global cluster assignments $\mathbf{A}$ from $P_g^T$
    \end{algorithmic}
\end{algorithm}

High-dimensional data often contains intricate structures that can impede effective clustering \cite{xie2023fed}. To enhance clustering performance, we integrate deep representation techniques that extract low-dimensional embeddings from high-dimensional data, thereby simplifying the data structure and improving cluster quality.

\begin{table*}[tb]
	\centering
    \setlength{\tabcolsep}{4pt}
	\begin{tabular}{c|c|ccccc|cc}
		\toprule
		\multirow{2}{*}{\textbf{Dataset}} & \multirow{2}{*}{\textbf{M}} & \multicolumn{5}{c|}{\textbf{Federated Methods}} & \multicolumn{2}{c}{\textbf{Our Methods}}\\ \cline{3-9} 
		& & \textbf{K-FED} & \textbf{FFCM} & \textbf{DSC} & \textbf{Fed-SC} & \textbf{PPFC-GAN} & \textbf{SPP-FGC} & \textbf{SPP-FGC+}\\
		\midrule
		\multirow{2}{*}{\textbf{Moon}} & A & 0.743$\pm$0.010 & 0.742$\pm$0.010 & 0.687$\pm$0.127 & 0.998$\pm$0.002 & 0.761$\pm$0.039 & \underline{0.987$\pm$0.039} & \textbf{0.999$\pm$0.001} \\
		& N & 0.178$\pm$0.015 & 0.177$\pm$0.016 & 0.265$\pm$0.025 & 0.930$\pm$0.118 & 0.215$\pm$0.062 & \underline{0.900$\pm$0.013} & \textbf{0.990$\pm$0.046}\\
		
		\multirow{2}{*}{\textbf{RING}} & A & 0.371$\pm$0.004 & 0.370$\pm$0.003 & 0.929$\pm$0.006 & 0.956$\pm$0.046 & 0.376$\pm$0.008 & \underline{0.971$\pm$0.018} & \textbf{0.990$\pm$0.011} \\
		& N & 0.433$\pm$0.002 & 0.433$\pm$0.002 & 0.858$\pm$0.049 & 0.830$\pm$0.000 & 0.432$\pm$0.003 & \underline{0.936$\pm$0.009} & \textbf{0.964$\pm$0.024}\\
		
		\multirow{2}{*}{\textbf{Letters}} & A & 0.149$\pm$0.011 & 0.250$\pm$0.003 & 0.281$\pm$0.000 & 0.319$\pm$0.019 & 0.260$\pm$0.009 & \underline{0.623$\pm$0.023} & \textbf{0.644$\pm$0.042} \\
		& N & 0.221$\pm$0.010 & 0.335$\pm$0.003 & 0.377$\pm$0.002 & 0.432$\pm$0.010 & 0.374$\pm$0.009 & \underline{0.563$\pm$0.015} & \textbf{0.614$\pm$0.013} \\

		\multirow{2}{*}{\textbf{MNIST}} & A & 0.559$\pm$0.559 & 0.490$\pm$0.004 & 0.602$\pm$0.001 & 0.612$\pm$0.027 & 0.518$\pm$0.013 & \underline{0.631$\pm$0.017} & \textbf{0.650$\pm$0.055} \\
		& N & 0.504$\pm$0.002 & 0.428$\pm$0.004 & 0.577$\pm$0.009 & 0.657$\pm$0.007 & 0.497$\pm$0.002 & \underline{0.719$\pm$0.011} & \textbf{0.740$\pm$0.011} \\
		
		\multirow{2}{*}{\textbf{Fashion}} & A & 0.518$\pm$0.005 & 0.601$\pm$0.001 & 0.503$\pm$0.006	& 0.600$\pm$0.013 & 0.568$\pm$0.002 & \underline{0.686$\pm$0.028} & \textbf{0.714$\pm$0.063} \\
		& N & 0.491$\pm$0.003 & 0.592$\pm$0.002 & 0.508$\pm$0.002 & 0.602$\pm$0.012 & 0.575$\pm$0.002 & \underline{0.670$\pm$0.036} & \textbf{0.709$\pm$0.027} \\
		
		\multirow{2}{*}{\textbf{CIFAR}}  & A & 0.725$\pm$0.000 & 0.656$\pm$0.003 & 0.610$\pm$0.001 & 0.705$\pm$0.016 & 0.726$\pm$0.005 & \underline{0.765$\pm$0.000} & \textbf{0.803$\pm$0.022} \\
		& N & 0.610$\pm$0.001 & 0.571$\pm$0.001 & 0.609$\pm$0.004 & 0.627$\pm$0.017 & 0.611$\pm$0.005 & \underline{0.646$\pm$0.001} & \textbf{0.683$\pm$0.031} \\
		
		\multirow{2}{*}{\textbf{STL-10}}  & A & 0.850$\pm$0.006 & 0.879$\pm$0.002 & 0.832$\pm$0.001 & 0.878$\pm$0.013 & 0.881$\pm$0.003 & \textbf{}{0.905$\pm$0.021} & \textbf{0.923$\pm$0.014} \\
		& N & 0.775$\pm$0.002 & 0.775$\pm$0.003 & 0.726$\pm$0.044 & 0.779$\pm$0.004 & 0.786$\pm$0.001 & \underline{0.810$\pm$0.033} & \textbf{0.865$\pm$0.015} \\
		
		\multirow{2}{*}{\textbf{M-Image}} & A & 0.733$\pm$0.010 & 0.438$\pm$0.010 & 0.399$\pm$0.028 & 0.369$\pm$0.002 & 0.448$\pm$0.009 & \underline{0.839$\pm$0.249} & \textbf{0.861$\pm$0.027} \\
		& N & 0.766$\pm$0.013 & 0.652$\pm$0.006 & 0.512$\pm$0.008 & 0.508$\pm$0.007 & 0.706$\pm$0.001 & \underline{0.810$\pm$0.037} & \textbf{0.850$\pm$0.017} \\
		\bottomrule
	\end{tabular}
	
	\caption{Clustering Performance Comparison on Real-World Datasets. The performance of SPP-FGC and SPP-FGC+ is evaluated against federated baselines using Clustering Accuracy (A) and Normalized Mutual Information (NMI). Each cell shows the mean ± standard deviation. Bold values denote the highest score in each row, while underlined values are the second-best.}
	\label{tab:comp}
\end{table*}

We propose the variant algorithm SPP-FGC+, which begins by using a VAE combined with a DEC loss to create effective embeddings for clustering. Specifically, the VAE is first pre-trained to learn robust initial embeddings and then fine-tuned with the DEC loss to optimize the latent space for clustering. The VAE encodes input features $X$ into latent embeddings $Z$, which are used to construct private structural graphs on each client. The total loss function $\mathcal{L}_{total}$ of the VAE-DEC framework is defined as:
\begin{equation}
	\mathcal{L}_{total} = \mathcal{L}_{vae} + \lambda \cdot \mathcal{L}_{dec}(P),
	\label{eq:16}
\end{equation}
where $\mathcal{L}_{vae}$ is the standard VAE reconstruction loss, and $\mathcal{L}_{dec}(P)$ is the clustering loss.

Each client builds a private structural matrix $E^{p,k}$ from the learned embeddings $Z$ and adds noise through DP to protect data during transmission. Protected prototypes are then sent to the central server, which combines them into a global similarity matrix and discover global structure as SPP-FGC. Finally, the global prototypes $P_g$ are sent back to all clients. These prototypes act as a common target for the next round of local learning. By optimizing its local VAE model to minimize the distance between its data embeddings and these common prototypes, each client is implicitly guided to align its private feature space with the global consensus established by the server.

This approach leverages the strengths of both self-supervised learning and FC to provide a robust solution for privacy-preserving, efficient clustering of complex data. The enhanced algorithm is summarized in Algorithm \ref{alg:SPP-FGC+}.

\section{Experimental Results}

We evaluated the SPP-FGC and SPP-FGC+ algorithms on both synthetic and real-world datasets with varying complexity. The synthetic datasets included Moon (two interleaving half circles) and Ring (a high-dimensional ring). Real-world datasets comprised simple tabular data (Iris, Breast Cancer, Bank, COLI20, USPS, and Letters) and more complex image data (MNIST, Fashion-MNIST, CIFAR-10, STL-10, and Mini-Imagenet), as summarized in Table 4 of Appendix F. For tabular data, all algorithms processed the raw input directly, while for images, SPP-FGC+ used a four-layer Variational Autoencoder (VAE), and other algorithms used 512-dimensional embeddings extracted via pretrained ResNet-18. All experiments were run in a CUDA 12.6 environment on an RTX 3090 GPU.

\subsection{Clustering Results Comparision}

We compared the performance of our SPP-FGC and SPP-FGC+ algorithms with other FC methods, including k-FED, FFCM, DSC, Fed-SC, and PPFC-GAN, to demonstrate their effectiveness in preserving data privacy and handling complex data scenarios. The experiments, conducted in a federated setting, revealed that our methods outperformed the baseline algorithms, in terms of Clustering Accuracy (Acc) and Normalized Mutual Information (NMI). In detail, clustering Acc is calculated after aligning predicted cluster labels with ground truth labels using the Hungarian algorithm to find the optimal one-to-one mapping. Notably, SPP-FGC+ achieved the highest NMI across most datasets, highlighting its superior ability to enhance clustering quality. Our methods excelled across eight selected datasets (with details in Appendix F), significantly outperforming other federated techniques. While SPP-FGC showed strong improvements on MNIST and Fashion-MNIST, its performance on CIFAR-10 and STL-10 suggested the potential for optimization in feature learning. SPP-FGC+'s adaptive feature embedding framework, which refines feature representations progressively, enabled more accurate clustering of high-dimensional and complex image datasets. In addition, the process of aggregation is shown in Figure.5 of Appendix F.I to visualize the effectiveness of clustering methods.

\subsection{Data Heterogeneous Experiment}
To evaluate the robustness of our framework against the critical challenge of data heterogeneity, we conducted an experiment on the MNIST dataset where the class imbalance across clients was systematically increased. As detailed in Table \ref{tab:Het}, SPP-FGC consistently outperforms baseline methods like Fed-Kmeans and Fed-SC across all levels of imbalance. While the performance of all algorithms naturally degrades as heterogeneity intensifies, the decline for SPP-FGC is far less severe, as it maintains significantly higher NMI and ACC scores. This demonstrates a remarkable resilience to the non-IID data distributions typical of practical federated applications, a conclusion further supported by visualizations in Appendix F.II.

\begin{table}[ht]
	\tabcolsep = 0.12cm
	\centering
	\begin{tabular}{c|cc|cc|cc}
		\toprule
		\multirow{2}{*}{\textbf{Het}} & \multicolumn{2}{c|}{\textbf{K-FED}} & \multicolumn{2}{c|}{\textbf{Fed-SC}} & \multicolumn{2}{c}{\textbf{SPP-FGC}} \\ \cline{2-7} 
		& NMI & ACC & NMI & ACC & NMI & ACC \\
		\hline
		0.2 & 0.4811 & 0.5474 & 0.6351 & 0.6113 & 0.6788 & 0.6224 \\
		0.4 & 0.4639 & 0.4707 & 0.5858 & 0.5109 & 0.5956 & 0.5403\\
		0.6 & 0.4495 & 0.4515 & 0.4751 & 0.4171 & 0.5476 & 0.5430\\
		0.8 & 0.4108 & 0.4415 & 0.2815 & 0.2203 & 0.5127 & 0.5073\\
		0.95 & 0.3864 & 0.3689 & 0.0279 & 0.1557 & 0.4849 & 0.4376\\
		\bottomrule
	\end{tabular}
        \caption{Robustness to Heterogeneity on the MNIST Dataset. The ``Het" column indicates the heterogeneity ratio, where a higher value signifies greater class imbalance across clients. }

	\label{tab:Het}
\end{table}

\subsection{Convergence Experiment}
\begin{figure}[t]
    \centering
    \begin{minipage}[b]{0.23\textwidth}
        \centering
        \includegraphics[width=\textwidth]{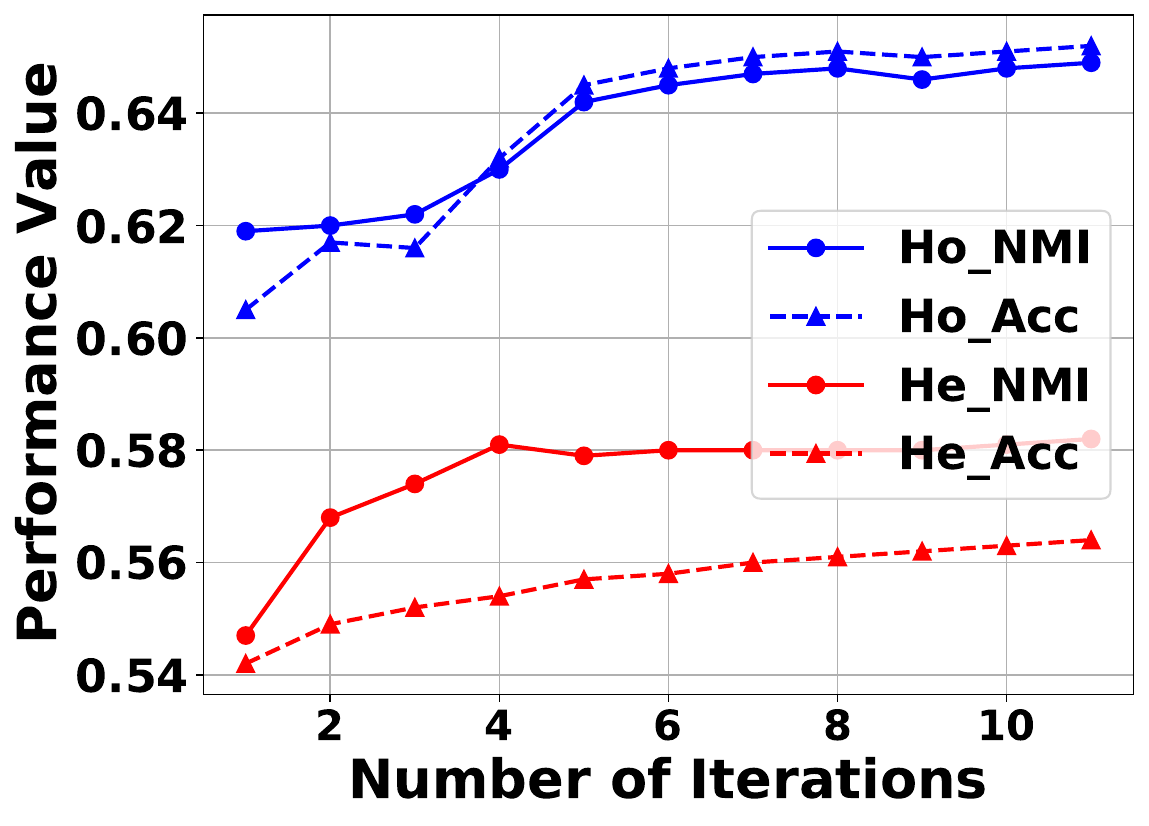}
        \label{fig:Conva}
    \end{minipage}
    \hfill
    \begin{minipage}[b]{0.23\textwidth}
        \centering
        \includegraphics[width=\textwidth]{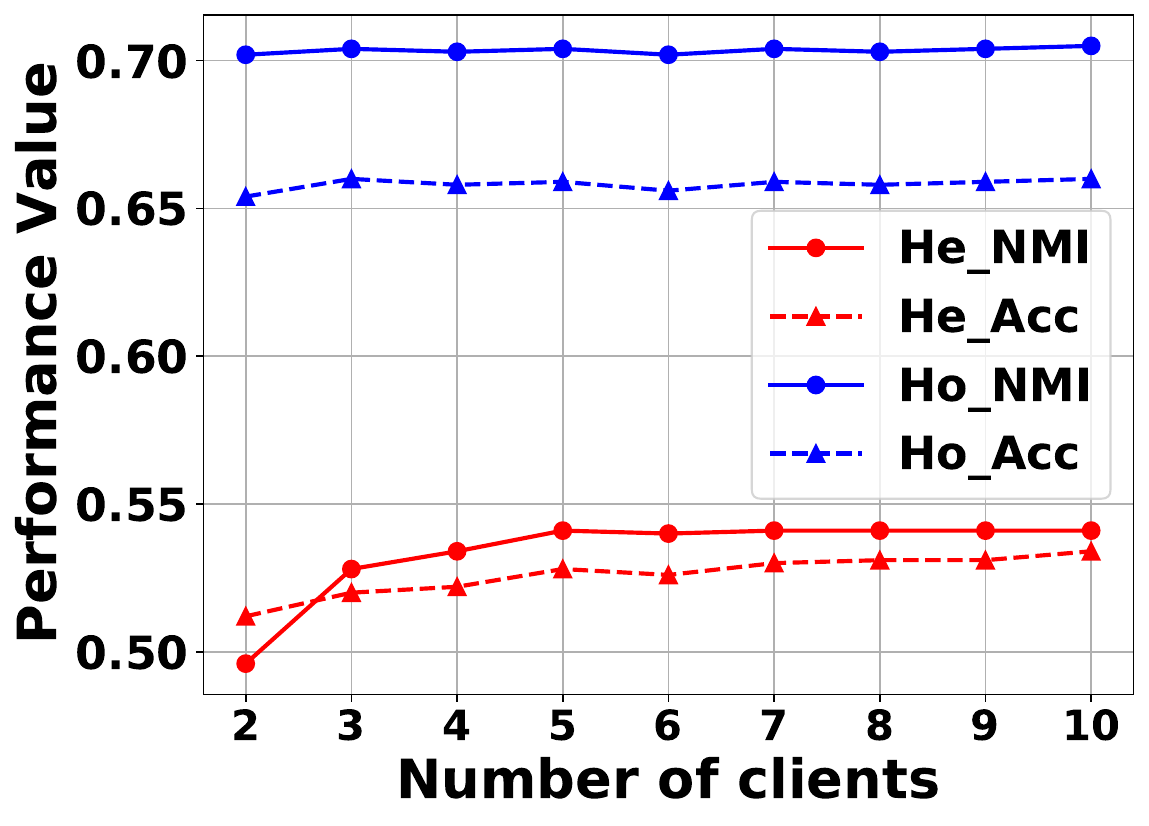}
        \label{fig:Convb}
    \end{minipage}
    \caption{Clustering performance measured by ACC and NMI. The Left: changes over the course of iterations, The Right: changes as the number of clients increases.}
    \label{fig:Conv}
\end {figure}

To validate the effectiveness of SPP-FGC+, we conducted two experiments on the MNIST dataset (shown in Fig \ref{fig:Conv}). The first tested performance scaling with the number of clients, showing improvement up to 10 clients, after which performance stabilized, demonstrating the method's scalability. The second examined convergence over training iterations, with performance steadily improving and plateauing, highlighting the benefits of iterative feature learning. These results confirm SPP-FGC+'s robustness and adaptability in large-scale federated environments and complex data scenarios. Detailed experiment setting and analysis are shown in Appendix F.III.

\subsection{Ablation Experiment}
    
	We conducted an ablation study on the SPP-FGC algorithm, testing five configurations: the original version, without Laplace Noise for privacy, without Private Structural Graph (PSG), without Global Structural Graph (GSG), and without both PSG and GSG. The results (Table \ref{tab:Abla}) show that while removing Laplace Noise slightly improved clustering performance, it compromised privacy, which is crucial for FL. Omitting either PSG or GSG caused noticeable performance drops, and removing both had the most significant impact. These findings confirm that PSG and GSG are essential for effective clustering, while Laplace Noise introduces a minor trade-off between privacy and performance.
	
    \begin{table}[h]
        \centering
        \begin{tabular}{ccc|c|c|c}
            \toprule
            \textbf{DP} & \textbf{PSG} & \textbf{GSG} &\textbf{NMI} & \textbf{ACC} & \textbf{ARI} \\
            \hline
            \Checkmark  & \Checkmark & \Checkmark & 0.7193 & 0.6305 & 0.6722 \\
            \XSolidBrush & \Checkmark & \Checkmark & 0.7228 & 0.6845 & 0.6931 \\
            \Checkmark & \XSolidBrush & \Checkmark & 0.5847 & 0.6256 & 0.5599 \\
            \Checkmark & \Checkmark & \XSolidBrush & 0.6793 & 0.6196 & 0.6417 \\
            \Checkmark & \XSolidBrush & \XSolidBrush & 0.5793 & 0.4720 & 0.4315 \\
            \bottomrule
        \end{tabular}
        \caption{Ablation Study of SPP-FGC. It assess the impact on NMI, ACC, and ARI when removing key components.}
        \label{tab:Abla}
    \end{table}
	
\subsection{Privacy Preservation Experiment}

\begin{figure}[h]
    \centering
    \begin{minipage}[b]{0.15\textwidth}
        \centering
        \includegraphics[width=\textwidth]{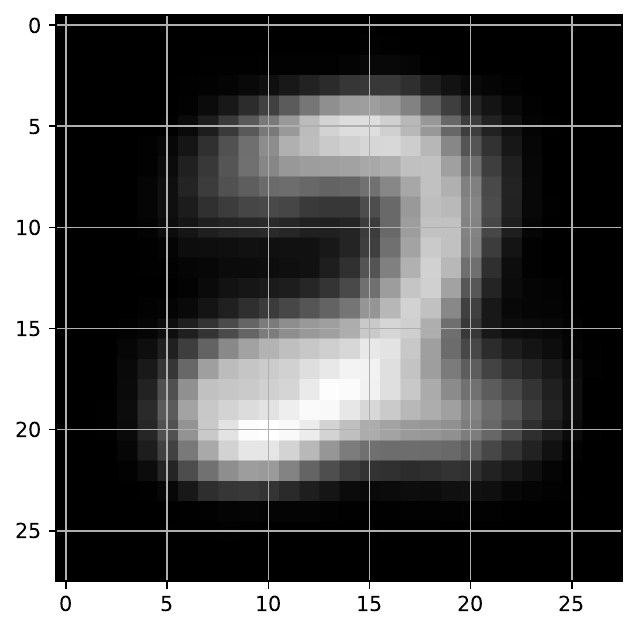}
        \label{fig:6a}
    \end{minipage}
    \hfill
    \begin{minipage}[b]{0.15\textwidth}
        \centering
        \includegraphics[width=\textwidth]{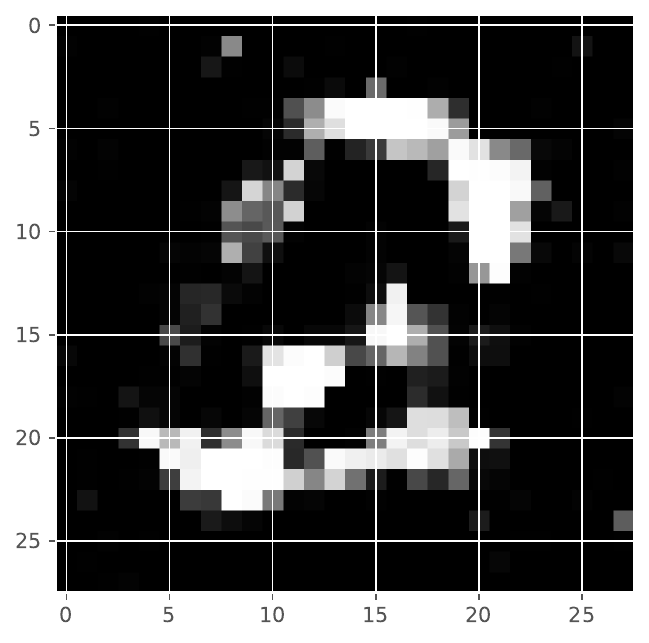}
        \label{fig:6b}
    \end{minipage}
    \hfill
    \begin{minipage}[b]{0.15\textwidth}
        \centering
        \includegraphics[width=\textwidth]{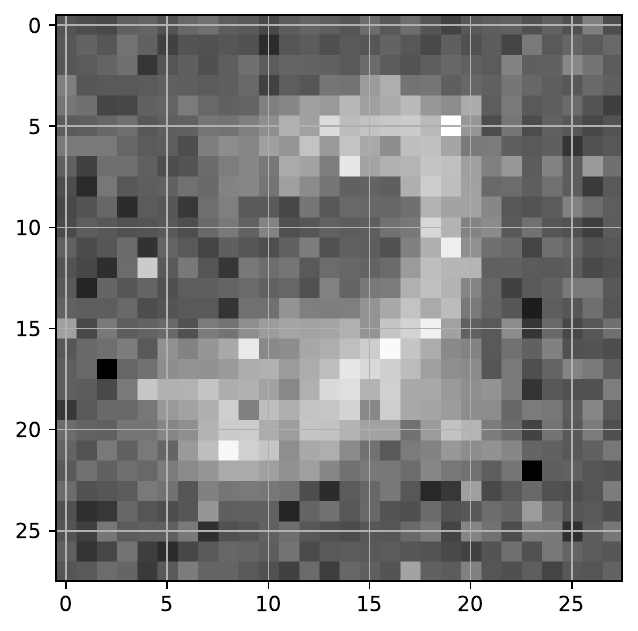}
        \label{fig:6c}
    \end{minipage}
    \begin{minipage}[b]{0.15\textwidth}
        \centering
        \includegraphics[width=\textwidth]{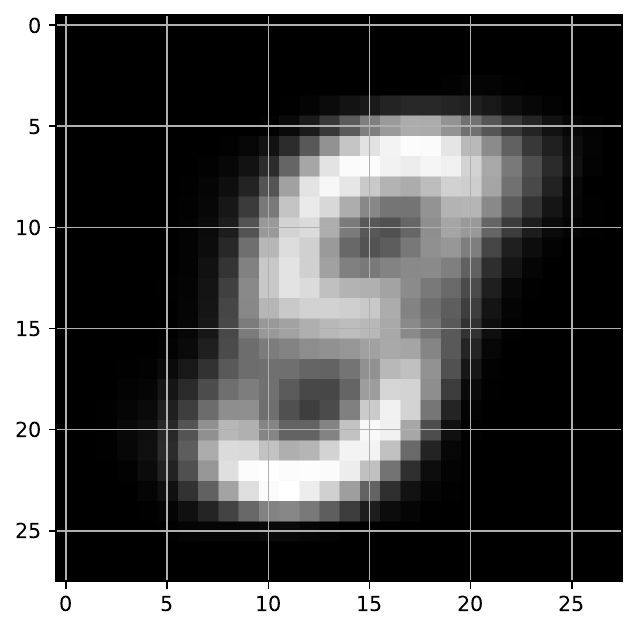}
        \label{fig:6d}
    \end{minipage}
    \hfill
    \begin{minipage}[b]{0.15\textwidth}
        \centering
        \includegraphics[width=\textwidth]{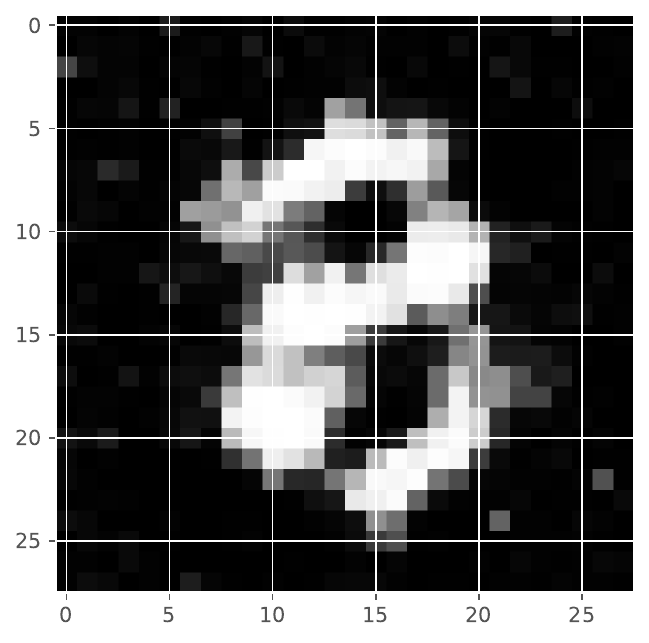}
        \label{fig:6e}
    \end{minipage}
    \hfill
    \begin{minipage}[b]{0.15\textwidth}
        \centering
        \includegraphics[width=\textwidth]{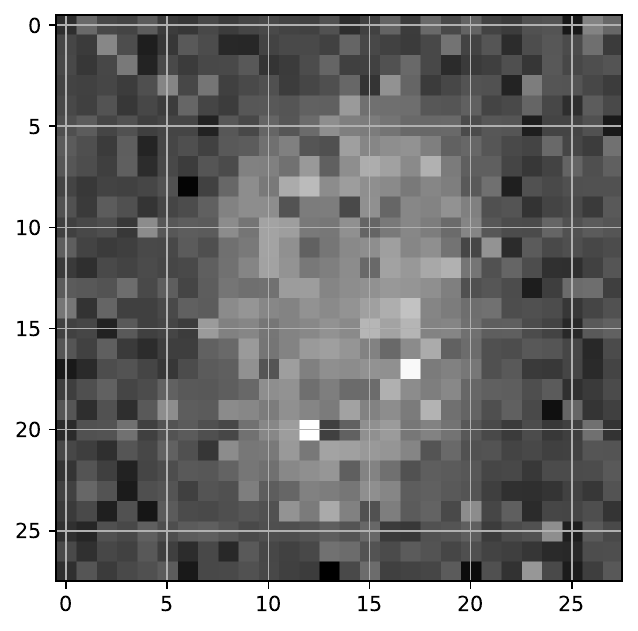}
        \label{fig:6f}
    \end{minipage}
    \caption{Visualization of transmitted prototypes from k-FED (left column), PPFC-GAN (middle column), and SPP-FGC (right column). This figure illustrates the enhanced privacy preservation achieved by our proposed SPP-FGC algorithm.}
    \label{fig:6}
\end {figure}

To demonstrate the privacy benefits of our DP method, we compared prototype visualizations from k-FED, PPFC-GAN, and our SPP-FGC on the MNIST dataset (on Fig \ref{fig:6}). While the prototypes from k-FED and PPFC-GAN closely resemble the original digits, potentially exposing sensitive data, SPP-FGC prototypes appear blurred, indicating stronger privacy protection. This highlights SPP-FGC's ability to safeguard individual data details while maintaining effective clustering, making it a reliable choice for privacy-sensitive FL applications.

    \section{Conclusion}
    In this paper, we propose SPP-FGC and SPP-FGC+, which enhance privacy and communication efficiency in FL through private structure mining, global relation aggregation, and DP. These approaches improve clustering performance while protecting sensitive data, making them suitable for privacy-sensitive applications. Our experiments demonstrated their effectiveness and scalability. Future work will focus on reducing costs with sparse graph representations and exploring adaptive privacy techniques to support diverse data distributions, further advancing FL's security and applicability.

    \section{Acknowledgments}
    This work was supported by the National Natural Science Foundation of China under Grant 62406250, 62236001, 62576277 and the Fundamental Research Funds for the Central Universities.
 
\bibliography{aaai2026}

    \newpage
    \appendix
    \setcounter{equation}{14}
    \setcounter{table}{3}
    \setcounter{figure}{4}
    \section{A. Privacy Structural Graph Learning}
    \label{APP:1}
    The original optimization target in Eq.3 is:
    \begin{equation*}
        \begin{aligned}
            &\underset{E^p}{\min} \sum_{i,j=1}^{N} \left( ||v_i - v_j||_2^2 E_{ij}^p + \gamma (E_{ij}^p)^2 \right), \\
            \text{s.t. } \forall i, \ &(E_{i}^p)^{\text{T}} \mathbf{1} = 1, \ 0 \leq E_{ij}^p \leq 1, \ \text{rank}(L_{E^p}) = N - C
        \end{aligned}
    \end{equation*}
    with the definition of Laplacian matrix in Eq.4:
    \begin{equation*}
        L_{E^p} = D_{E^p} - \frac{(E^p)^{\text{T}} + E^p}{2}.
    \end{equation*}
    
    According to the definition of the Laplacian matrix, $L_{E^p}$ is positive semidefinite, meaning the $i$-th smallest eigenvalue of the matrix satisfies $\sigma_i(L_{E^p}) \geq 0$. Therefore, the problem in Eq.3 can be converted to:
    
    \begin{equation}
        \begin{aligned}
            \underset{E^p}{\min} \sum_{i,j=1}^{N} &\left( ||v_i - v_j||_2^2 E_{ij}^p + \gamma (E_{ij}^p)^2 \right) + 2\lambda \sum_{i=1}^C \sigma_i(L_{E^p}), \\
            &\text{s.t. } \forall i, \ (E_{i}^p)^{\text{T}} \mathbf{1} = 1, \ 0 \leq E_{ij}^p \leq 1
        \end{aligned}
    \end{equation}
    with a sufficiently large $\lambda$, the term $\sum_{i=1}^C \sigma_i(L_{E^p})$ is forced to zero, thereby satisfying the constraint $\text{rank}(L_{E^p}) = N - C$. By defining $F \in \mathbb{R}^{N \times C}$, Ky Fan’s Theorem proves that:
    
    \begin{equation}
        \sum_{i=1}^{C} \sigma_i(L_{E^p}) = \underset{F^{\text{T}}F = I}{\min} \text{Tr}(F^{\text{T}} L_{E^p} F),
    \end{equation}
    which transforms problem into:
    
    \begin{equation}
        \begin{aligned}
            \underset{E^p,F}{\min} \sum_{i,j=1}^{N} \left( ||v_i - v_j||_2^2 E_{ij}^p + \gamma (E_{ij}^p)^2 \right) &+ 2\lambda \text{Tr}(F^{\text{T}} L_{E^p} F). \\
            \text{s.t. } \forall i, \ (E_{i}^p)^{\text{T}} \mathbf{1} = 1, \ 0 \leq E_{ij}^p &\leq 1, \ F^{\text{T}} F = I
        \end{aligned}
    \end{equation}
    
    This problem is easier to solve than the original one. By employing the alternating optimization method, we can simplify the optimization process by fixing one of the matrices $F$ or $E^p$ and solving for the other.
    
    With $E^p$ fixed, problem Eq.7 simplifies to:
    
    \begin{equation}
        \begin{aligned}
            \underset{F}{\min} \ \text{Tr}&(F^{\text{T}} L_{E^p} F). \\
            \text{s.t. } \ &F^{\text{T}} F = I
        \end{aligned}
    \end{equation}
    
    The best solution to this problem is the matrix constructed by the $C$ eigenvectors of $L_{E^p}$ corresponding to the $C$ smallest eigenvalues.
    
    When $F$ is fixed, a new optimization problem arises:
    
    \begin{equation}
        \begin{aligned}
            \underset{E_i^p}{\min} \sum_{j=1}^{N} ( ||v_i &- v_j||_2^2 E_{ij}^p + \gamma (E_{ij}^p)^2 + \lambda ||f_i - f_j||_2^2 ). \\
            &\text{s.t. } \ (E_{i}^p)^{\text{T}} \mathbf{1} = 1, \ 0 \leq E_{ij}^p \leq 1
        \end{aligned}
    \end{equation}
    
    This problem can be converted into vector form as:
    
    \begin{equation}
        \begin{aligned}
            &\underset{E_i^p}{\min} \ ||E_i^p + \frac{d_i}{2\gamma}||_2^2, \\
            \text{s.t. } \ &(E_i^p \mathbf{1} = 1, \ 0 \leq E_i^p \leq 1)
        \end{aligned}
    \end{equation}
    where $d_{ij}^x = ||v_i - v_j||_2^2$, $d_{ij}^f = ||f_i - f_j||_2^2$, and the vector $d_i$ has elements $d_{ij} = d_{ij}^x + d_{ij}^f$.
    
    The Lagrangian function for each $i$ is:
    
    \begin{equation}
        \mathcal{L}(E_i^p, \eta, \beta_i) = ||E_i^p + \frac{d_i}{2\gamma_i}||_2^2 - \eta \left( (E_i^p)^{\text{T}} \mathbf{1} - 1 \right) - \beta_i^{\text{T}} E_i^p,
    \end{equation}
    where $\eta$ and $\beta_i \geq 0$ are Lagrangian multipliers. Using the Karush-Kuhn-Tucker (KKT) conditions, the optimal solution $E_i^p$ is:
    
    \begin{equation}
        E_{ij}^p = -\frac{d_{ij}}{2\gamma_i} + \eta.
    \end{equation}
    
    Assuming only the $k$ nearest neighbors are considered, where $d_{i1}, d_{i2}, \dots, d_{iN}$ are ordered from smallest to largest, with only $k$ nonzero elements, the following conditions hold:
    
    \begin{equation}
        \left\{
        \begin{aligned}
            -\frac{d_{ik}}{2\gamma_i} + \eta &> 0, \\
            -\frac{d_{i,k+1}}{2\gamma_i} + \eta &\leq 0,
        \end{aligned}
        \right.
    \end{equation}
    
    Given $E_i^p \mathbf{1} = 1$, we obtain:
    
    \begin{equation}
        \sum_{j=1}^{k} \left( -\frac{d_{ij}}{2\gamma_i} + \eta \right) = 1,
    \end{equation}
    leading to:
    \begin{equation}
        \eta = \frac{1}{k} + \frac{1}{2k\gamma_i} \sum_{j=1}^{k} d_{ij}.
    \end{equation}
    
    This implies that $\gamma_i$ satisfies the inequality:
    
    \begin{equation}
        \frac{k}{2} d_{i,k+1} - \frac{1}{2} \sum_{j=1}^{k} d_{ij} < \gamma_i \leq \frac{k}{2} d_{ik} - \frac{1}{2} \sum_{j=1}^{k} d_{ij}.
    \end{equation}
    By setting $\gamma_i = \frac{k}{2} d_{i,k+1} - \frac{1}{2} \sum_{j=1}^{k} d_{ij}$, the average $\gamma$ across all samples is:
    
    \begin{equation}
        \gamma = \frac{1}{N} \sum_{i=1}^{N} \left( \frac{k}{2} d_{i,k+1} - \frac{1}{2} \sum_{j=1}^{k} d_{ij} \right).
    \end{equation}
    
    Thus, with a fixed number of neighbors $k$, the private structural matrix $E^p$ has a closed-form solution as shown in Eq.11.

    \section{B. Global Structural Graph Learning}
    \label{APP:2}
    For the optimization problem of Eq.11:
    \begin{equation*}
        \begin{aligned}
            &\underset{S}{\min} \ ||S - E^*||_F^2. \\
            \text{s.t.} \quad& \text{rank}(L_S) = N - C
        \end{aligned}
        \label{eq:23}
    \end{equation*}
    This problem can be reformulated as:
    
    \begin{equation}
        \begin{aligned}
            \underset{S}{\min} \ ||S - E^*||_F^2& + 2\lambda \sum_{i=1}^C \sigma_i(L_S).\\
            \text{s.t.} \quad S_{ij} \geq 0, &\sum_j S_{ij} = 1
        \end{aligned}
        \label{eq:24}
    \end{equation}
    Applying Ky Fan’s Theorem, we transfer it to:
    
    \begin{equation}
        \begin{aligned}
            &\underset{S}{\min} \ ||S - E^*||_F^2 + 2\lambda \text{Tr}(F^{\text{T}} L_S F), \\
            \text{s.t.} \quad \sum_j &S_{ij} = 1, \ S_{ij} \geq 0, \ F \in \mathbb{R}^{N \times C}, \ F^{\text{T}} F = I
        \end{aligned}
        \label{eq:25}
    \end{equation}
    
    Similarly, we employ an alternating optimization approach to solve problem Eq.\ref{eq:25}. This involves iteratively optimizing $F$ and $S$ by fixing one and solving for the other.
    
    When $S$ is fixed, the optimization problem becomes:
    
    \begin{equation}
        \begin{aligned}
            &\underset{F}{\min} \ \text{Tr}(F^{\text{T}} L_S F). \\
            \text{s.t.} \quad &F \in \mathbb{R}^{N \times C}, \ F^{\text{T}} F = I
        \end{aligned}
        \label{eq:26}
    \end{equation}
    The optimal solution to this problem is obtained by selecting the $C$ eigenvectors of $L_S$ corresponding to the $C$ smallest eigenvalues.
    
    Conversely, when $F$ is fixed, the optimization problem for $S$ becomes:
    
    \begin{equation}
        \begin{aligned}
            \underset{S}{\min} \ \sum_{i,j} &\left( S_{ij} - E_{ij}^* \right)^2 + \lambda \sum_{i,j} ||f_i - f_j||_2^2 S_{ij}. \\
            &\text{s.t.} \quad \sum_j S_{ij} = 1, \ S_{ij} \geq 0
        \end{aligned}
        \label{eq:27}
    \end{equation}
    
    This problem can be decomposed for each $i$ as:
    
    \begin{equation}
        \begin{aligned}
            \min_{S_{ij}} \ \sum_{j} &\left( S_{ij} - E_{ij}^* \right)^2 + \lambda \sum_{j} ||f_i - f_j||_2^2 S_{ij}. \\
            &\text{s.t.} \quad \sum_j S_{ij} = 1, \ S_{ij} \geq 0
        \end{aligned}
        \label{eq:28}
    \end{equation}
    By defining $d^f_{ij} = ||f_i - f_j||_2^2$, $d^v_{ij} = ||v_i - v_j||_2^2$, and $d_{ij} = d^v_{ij} + d^f_{ij}$, the problem in Eq.\ref{eq:28} can be reformulated as:
    
    \begin{equation}
        \begin{aligned}
            \min_{S_i} \ ||S_i - E_i^* + \frac{\lambda}{2} d_i^f||_2^2, \\
            \text{s.t.} \quad \sum_j S_{ij} = 1, \ S_{ij} \geq 0
        \end{aligned}
        \label{eq:29}
    \end{equation}
    
    The Lagrangian function for each $i$ is:
    
    \begin{equation}
        \mathcal{L}(S_i, \eta, \beta_i) = ||S_i - E_i^* + \frac{\lambda}{2} d_i^f||_2^2 - \eta \left( \sum_j S_{ij} - 1 \right) - \beta_i^{\text{T}} S_i,
        \label{eq:30}
    \end{equation}
    where $\eta$ and $\beta_i \geq 0$ are Lagrangian multipliers. Applying the Karush-Kuhn-Tucker (KKT) conditions, the optimal solution $S_i$ is:
    
    \begin{equation}
        S_{ij} = -\frac{d_{ij}}{2\gamma} + \eta,
        \label{eq:31}
    \end{equation}
    where $\gamma$ is derived based on the constraints. By iteratively refining $S$, the clustering result can be directly obtained from its clear block diagonal structure or further enhanced using a simple k-means method. These clustering centers can then be disseminated to all clients, providing global class prototypes that aid local clients in making more informed decisions from a global perspective.
    
    \section{C. Privacy Guarantee via Differential Privacy}
    \subsection{C.I $\epsilon$-DP Proof}
    
    \textbf{Theorem:} The transmission of prototypes $P_k=\{\mu_{k,c},\Sigma_{k,c}\}_{c=1}^C$ using the Laplace mechanism as described above satisfies $\epsilon$-Differential Privacy.
    
    \emph{Proof:} The Laplace mechanism ensures $\epsilon$-DP by adding noise calibrated to the sensitivity of the function being computed. Specifically, for a function $f$ with sensitivity $\Delta_f$, adding Laplace noise $Lap\left(\Delta_f/\epsilon\right)$ to each component of $f(D)$ guarantees that the mechanism $\mathcal{M}(D) = f(D) + \text{Lap}\left(\Delta_f/\epsilon\right)$ satisfies $\epsilon$-DP.
    
    For the prototypes $\mu_k$ and $\Sigma_k$, sensitivities $\Delta_{\mu}$ and $\Delta_{\Sigma}$ are defined based on the maximum changes in these parameters due to the alteration of a single data point. Adding Laplace noise $\text{Lap}\left(\Delta_{\mu}/\epsilon\right)$ and $\text{Lap}\left(\Delta_{\Sigma}/\epsilon\right)$ to each element of $\mu_k$ and $\Sigma_k$, respectively, ensures $\epsilon$-DP for the prototypes.
    
    Since the noise addition for $\mu_i$, and $\Sigma_i$ is independent and calibrated according to their respective sensitivities, the overall mechanism satisfies $\epsilon$-Differential Privacy by the composition properties of DP. Formally, for all datasets $D$ and $D'$ differing by a single element, and for all measurable subsets $S$ of the output space, the following holds:
    \begin{equation}
        \Pr[\mathcal{M}(D) \in S] \leq e^{\epsilon} \Pr[\mathcal{M}(D') \in S],
    \end{equation}
    
    which is the definition of $\epsilon$-Differential Privacy, ensuring that the probability of any particular output $S$ does not change by more than a factor of $e^{\epsilon}$ when any single individual's data is added or removed from the dataset. Consequently, this guarantees that the presence or absence of any individual data point has a negligible impact on the output, thereby protecting the privacy of each participant in the federated learning process.
    
    \subsection{C.II Analysis on Real Data}
    
    The Laplace noise added to prototypes $\mu_k$ and $\sigma_k$ with sensitivities $\Delta_{\mu}$ and $\Delta_{\Sigma}$. With normalized data $\|x_i\|_1 = 1$, we can calculate the sensitivities mathematically.
    
    \textbf{Sensitivity of Mean prototype}
    \begin{align*}
        \Delta_{\mu} &= \max_{D,D'} \| \mu_{k,c}(D) - \mu_{k,c}(D') \|_1 \\
        &= \max_{x_j,x'_j} \left\| \frac{1}{N_c}(x'_j - x_j) \right\|_1 \\
        &\leq \frac{1}{N_{c_{min}}} \max_{x_j,x'_j} \|x'_j - x_j\|_1 \\
        &= \frac{2}{N_{c_{min}}}
    \end{align*}
    
    \textbf{Sensitivity of Covariance Prototype}
    \begin{align*}
        \Delta_{\Sigma} &= \max_{D,D'} \|\Sigma_{k,c}(D) - \Sigma_{k,c}(D')\|_F \\
        &\leq \frac{1}{N_{c_{min}}} \Big( \|x'_j x_j^{\prime\top} - x_j x_j^\top\|_F \\
        &\quad + 2\|\mu_{k,c}(x'_j - x_j)^\top\|_F \Big) \\
        &\leq \frac{2\sqrt{2} + 4}{N_{c_{min}}}
    \end{align*}
    
    \subsection{C.III Conclusion}
    The calibrated Laplace mechanism preserves critical population statistics while obscuring individual contributions. By exploiting the \textbf{concentration effect} of prototypes ($\mu_k,\Sigma_k$) and the \textbf{self-normalization} property of L1-constrained data, the method achieves:
    \begin{itemize}
        \item Provable privacy guarantees through bounded sensitivities
        \item Maintained utility via bias-variance tradeoff control
        \item Practical efficiency through dimension-independent noise scaling
    \end{itemize}
    
    This enables secure knowledge sharing in federated learning while preserving the geometric structure of feature representations.
    
    \section{D. Privacy Guarantee via Information Entropy}
    
    In FL frameworks, safeguarding data privacy during transmission from clients to the central server is crucial. In our SPP-FGC+ algorithm, clients transmit their private structural matrixs $E^p$ and class prototypes $\{\mu_{k,c}, \Sigma_{k,c}\}_{c=1}^C$ instead of the raw feature matrices $X$. This section mathematically demonstrates that transmitting $E^p$ results in lower information entropy compared to transmitting $X$, thereby enhancing privacy protection.
    
    Information entropy $H(\cdot)$ quantifies the uncertainty or information content of a random variable. For a continuous random variable $Y$ with probability density function $p(y)$, the differential entropy is defined as:
    \begin{equation}
        H(Y) = -\int p(y) \log p(y) \, dy
    \end{equation}
    
    Consider the raw feature matrix $X \in \mathbb{R}^{n \times d}$, where $n$ is the number of samples and $d$ is the feature dimensionality. Assuming each feature vector $v_i$ ($i=1,\dots,n$) is independently and identically distributed (i.i.d.), the entropy of $X$ is:
    \begin{equation}
        H(X) = \sum_{i=1}^{n} H(v_i)
    \end{equation}
    In scenarios involving complex unstructured data or a high number of clients, $d$ typically satisfies $d \gg n$, resulting in a substantial $H(X)$ due to high dimensionality.
    
    The private structural matrix $E^p \in \mathbb{R}^{n \times n}$ represents pairwise similarities between samples based on latent embeddings $Z$. $E^p$ has the following properties:
    \begin{enumerate}
        \item Each row sums to 1: $\sum_{j=1}^{n} E^p_{ij} = 1$, making $E^p$ a row-stochastic matrix.
        \item The Laplacian matrix $L_{E^p}$ has rank $n - C$, where $C$ is the number of clusters.
    \end{enumerate}
    Given that $E^p$ is structured as a block diagonal matrix with $C$ blocks, the entropy can be approximated as:
    \begin{equation}
        H(E^p) \approx \sum_{i=1}^{n} \sum_{j=1}^{n} H(E^p_{ij})
    \end{equation}
    
    To compare $H(E^p)$ and $H(X)$, consider:
    \begin{enumerate}
        \item \textbf{Block Diagonal Sparsity:}  
        The block diagonal structure of $E^p$ with $C$ blocks means each row has at most $k$ non-zero entries ($k \ll n$), significantly reducing the number of high-entropy elements compared to the dense $X$.
        
        \item \textbf{Noise Calibration for DP:}  
        Noise added to $E^p$ for DP is calibrated based on its lower sensitivity due to sparsity, maintaining lower entropy compared to noise added to the high-dimensional $X$.
    \end{enumerate}
    
    Mathematically, assuming $X \sim \mathcal{N}(0, \Sigma_X)$ and $E^p \sim \mathcal{N}(0, \Sigma_{E^p})$ with $\Sigma_{E^p}$ being significantly sparser than $\Sigma_X$, the entropies are:
    \begin{equation}
        H(X) = \frac{1}{2} \log\left((2\pi e)^d |\Sigma_X|\right)
    \end{equation}
    \begin{equation}
        H(E^p) = \frac{1}{2} \log\left((2\pi e)^{k} |\Sigma_{E^p}|\right)
    \end{equation}
    Since $|\Sigma_{E^p}| \leq |\Sigma_X|$ due to sparsity and $k \ll d$, it follows that:
    \begin{equation}
        H(E^p) \leq H(X)
    \end{equation}
    
    Transmitting $E^p$ instead of $X$ enhances privacy through:
    
    \begin{enumerate}
        \item \textbf{Lower Entropy Limits Reconstruction:}  
        The private structural matrix has lower entropy than the raw feature matrix ($H(E^p) \leq H(X)$). This makes it harder for adversaries to accurately reconstruct individual data points from $E^p$, as it contains less information about the original data.
        
        \item \textbf{Efficient Use of Privacy Budget:}  
        Lower entropy in $E^p$ allows for more effective utilization of the privacy budget $\epsilon$, balancing privacy and utility more efficiently than with high-entropy data like $X$.
    \end{enumerate}

    Transmitting the private structural matrix $E^p$ in SPP-FGC and SPP-FGC+ reduces information entropy compared to sending the raw feature matrix $X$. This reduction is achieved through dimensionality reduction in scenarios with complex data or many clients, the sparse block diagonal structure of $E^p$, and controlled noise addition via Differential Privacy. As a result, our methods provides enhanced privacy protection, making it a more secure and efficient method for federated clustering of high-dimensional data.

    \section{E. Complexity Analysis and Sparsity}
    
    Our framework is designed for efficiency, leveraging sparse graph structures to reduce both computational and communication overhead.

    On the client side, instead of constructing a dense graph with quadratic complexity ($O(N_k^2)$), we build a sparse \textbf{$k_n$-NN graph}. This is highly efficient, as finding neighbors can be optimized to be much faster than a quadratic operation.
    
    This client-side sparsity is the key to server-side efficiency. The server aggregates these sparse graphs into a global graph, which itself remains sparse. A naive approach operating on a dense $N \times N$ matrix would incur prohibitive quadratic or cubic costs. However, by exploiting the sparse nature of the global graph, the server can utilize efficient iterative eigensolvers. This is the crucial step that reduces the server's complexity from quadratic to near-linear ($O(Nk_n)$), where $N$ is the total number of samples and $k_a$ is the small average number of neighbors.
    
    Finally, the communication cost is also simplified to near-linear. Clients do not transmit dense matrices or high-dimensional embeddings for every sample. Instead, they send only the non-zero entries of their sparse $k_n$-NN graphs ($O(N_k k)$ values) and a small, constant-sized set of cluster prototypes. This makes the total communication cost proportional to the number of samples, not the square of the samples, ensuring scalability in a federated setting.

    \section{F. Experimental Analysis}
    \label{APP:Exp}
    \begin{table}[!htb]
        \centering
        \caption{Overview of used datasets.}
        \begin{tabular}{cccc}
            \hline
            \textbf{Dataset} & \textbf{\# Classes} & \textbf{\# Dimensions} & \textbf{\# Samples} \\ \hline
            Iris & 3 & 4 & 150 \\
            Cancer & 2 & 30 & 569 \\
            Moon & 2 & 2 & 1000\\
            Bank & 2 & 4 & 1,372 \\
            COIL20 & 20 & 1024 & 1440 \\
            RING & 5 & 20 & 5000\\
            USPS & 10 & 256 & 9,298 \\
            Letters & 26 & 16 & 20,000 \\
            MNIST & 10 & 28*28 & 70,000 \\
            Fashion & 10 & 28*28 & 70,000 \\
            CIFAR-10 & 10 & 3*32*32 & 60,000 \\
            STL-10 & 10 & 3*96*96 & 13,000 \\ 
            Mini-Imagenet & 100 & 3*84*84 &6,0000 \\
            \hline
        \end{tabular}
        \label{tab:data}
    \end{table}
    
    \subsection{F.I Comparision Experiment Details}
    \begin{table*}[!htb]
        \tabcolsep = 0.1cm
        \centering
        \caption{Clustering performance comparison for FC methods using 10 datasets.}
        \resizebox{\textwidth}{!}{\begin{tabular}{c|c|ccccc|cc}
            \hline
            \multirow{2}{*}{\textbf{Dataset}} & \multirow{2}{*}{\textbf{M}} & \multicolumn{5}{c|}{\textbf{Federated Methods}} & \multicolumn{2}{c}{\textbf{Our Methods}}\\ \cline{3-9} 
            & & \textbf{k-FED} & \textbf{FFCM} & \textbf{DSC} & \textbf{Fed-SC} & \textbf{PPFC-GAN} & \textbf{SPP-FGC} & \textbf{SPP-FGC+}\\ \hline
            
            \multirow{2}{*}{\textbf{Iris}}& A & 0.6600$\pm$0.0163 & 0.8933$\pm$0.0217 & 0.8460$\pm$0.0142
            & 0.8960$\pm$0.0076 & 0.8867$\pm$0.0079 & \textbf{0.9629$\pm$0.0138} & \textbf{0.9688$\pm$0.0216} \\
            & N & 0.6536$\pm$0.0304 & 0.7398$\pm$0.0303 & 0.6787$\pm$0.0726 & 0.7731$\pm$0.0210 & 0.7419$\pm$0.0032 & \textbf{0.8818$\pm$0.0336} & \textbf{0.8922$\pm$0.0317}\\ \hline
            
            \multirow{2}{*}{\textbf{Cancer} } & A & 0.8500$\pm$0.0160 & 0.8318$\pm$0.0181 & 0.8362$\pm$0.0047 & 0.8243$\pm$0,0030 & 0.8187$\pm$0.0322 & \textbf{0.9425$\pm$0.0143} & \textbf{0.9501$\pm$0.0319} \\  
            & N & 0.6472$\pm$0.0300 & 0.4310$\pm$0.0391 & 0.4712$\pm$0.0645 & 0.6342$\pm$0.0051 & 0.4218$\pm$0.0679 & \textbf{0.6977$\pm$0.0245} & \textbf{0.7102$\pm$0.0319} \\ \hline
            
            \multirow{2}{*}{\textbf{Moon}} & A & 0.7426$\pm$0.0099 & 0.7420$\pm$0.0101 & 0.6870$\pm$0.1270 & 0.9983$\pm$0.0020 & 0.7610$\pm$0.0385 & \textbf{0.9866$\pm$0.0393} & \textbf{0.9990$\pm$0.0005} \\
            & N & 0.1780$\pm$0.0153 & 0.1770$\pm$0.0156 & 0.2654$\pm$0.0249 & 0.9297$\pm$0.1175 & 0.2148$\pm$0.0617 & \textbf{0.9003$\pm$0.0131} & \textbf{0.9895$\pm$0.0455}\\ \hline
            
            \multirow{2}{*}{\textbf{Bank}} & A & 0.5113$\pm$0.0078 & 0.6171$\pm$0.0116 & 0.5624$\pm$0.0042 & 0.8402$\pm$0.0636 & 0.6027$\pm$0.0018 & \textbf{0.9333$\pm$0.0092} & \textbf{0.9505$\pm$0.0118} \\
            & N & 0.0241$\pm$0.0078 & 0.0347$\pm$0.0064 & 0.0282$\pm$0.0169 & 0.6273$\pm$0.1397 & 0.0396$\pm$0.0021 & \textbf{0.7093$\pm$0.0275} & \textbf{0.7155$\pm$0.0219} \\ \hline
            
            \multirow{2}{*}{\textbf{COIL}} & A & 0.4703$\pm$0.0274 & 0.5675$\pm$0.0078 & 0.4590$\pm$0.0007 & 0.8023$\pm$0.0011 & 0.4438$\pm$0.0084 & \textbf{0.8883$\pm$0.0127} & \textbf{0.9007$\pm$0.0251} \\
            & N & 0.6896$\pm$0.0040 & 0.7595$\pm$0.0078 & 0.6952$\pm$0.0017 & 0.8830$\pm$0.0012 & 0.5638$\pm$0.0035 & \textbf{0.9072$\pm$0.0169} & \textbf{0.9122$\pm$0.0313} \\ \hline
            
            \multirow{2}{*}{\textbf{RING}} & A & 0.3706$\pm$0.0039 & 0.3697$\pm$0.0026 & 0.9292$\pm$0.0062 & 0.9564$\pm$0.0462 & 0.3758$\pm$0.0084 & \textbf{0.9710$\pm$0.0183} & \textbf{0.9895$\pm$0.0106} \\
            & N & 0.4330$\pm$0.0021 & 0.4327$\pm$0.0021 & 0.8577$\pm$0.0492 & 0.8295$\pm$0.0001 & 0.4317$\pm$0.0026 & \textbf{0.9359$\pm$0.0085} & \textbf{0.9642$\pm$0.0243}\\ \hline
            
            \multirow{2}{*}{\textbf{USPS}} & A & 0.5990$\pm$0.0103 & 0.6269$\pm$0.0218 & 0.5627$\pm$0.0012 & 0.6632$\pm$0.0016 & 0.7261$\pm$0.0047& \textbf{0.8681$\pm$0.0723} & \textbf{0.8977$\pm$0.0723} \\
            & N & 0.5839$\pm$0.0060 & 0.5746$\pm$0.0147 & 0.5331$\pm$0.0024 & 0.6632$\pm$0.0012 & 0.6108$\pm$0.0051 & \textbf{0.8624$\pm$0.0473} & \textbf{0.8891$\pm$0.0027} \\ \hline
            
            \multirow{2}{*}{\textbf{Letters}} & A & 0.1489$\pm$0.0105 & 0.2498$\pm$0.0034 & 0.2812$\pm$0.0004 & 0.3194$\pm$0.0194 & 0.2599$\pm$0.0085 & \textbf{0.6225$\pm$0.0227} & \textbf{0.6443$\pm$0.0417} \\
            & N & 0.2207$\pm$0.0102 & 0.3346$\pm$0.0034 & 0.3765$\pm$0.0016 & 0.4316$\pm$0.0097 & 0.3738$\pm$0.0088 & \textbf{0.5631$\pm$0.0150} & \textbf{0.6144$\pm$0.0131} \\ \hline
            
            \multirow{2}{*}{\textbf{MNIST}} & A & 0.5586$\pm$0.5586 & 0.4902$\pm$0.0044 & 0.6021$\pm$0.0005 & 0.6115$\pm$0.0266 & 0.5175$\pm$0.0126 & \textbf{0.6305$\pm$0.0171} & \textbf{0.6504$\pm$0.0547} \\
            & N & 0.5041$\pm$0.0020 & 0.4282$\pm$0.0043 & 0.5772$\pm$0.0088 & 0.6569$\pm$0.0066 & 0.4967$\pm$0.0021 & \textbf{0.7193$\pm$0.0106} & \textbf{0.7402$\pm$0.0114} \\ \hline
            
            \multirow{2}{*}{\textbf{Fashion}} & A & 0.5178$\pm$0.0053 & 0.6008$\pm$0.0014 & 0.5026$\pm$0.0057	& 0.5996$\pm$0.0131 & 0.5680$\pm$0.0024 & \textbf{0.6856$\pm$0.0275} & \textbf{0.7139$\pm$0.0628} \\
            & N & 0.4914$\pm$0.0034 & 0.5915$\pm$0.0022 & 0.5079$\pm$0.0016 & 0.6023$\pm$0.0116 & 0.5751$\pm$0.0018 & \textbf{0.6699$\pm$0.0358} & \textbf{0.7093$\pm$0.0272} \\ \hline
            
            \multirow{2}{*}{\textbf{CIFAR}}  & A & 0.7252$\pm$0.0001 & 0.6560$\pm$0.0026 & 0.6102$\pm$0.0013 & 0.7048$\pm$0.0164 & 0.7261$\pm$0.0047 & \textbf{0.7651$\pm$0.0002} & \textbf{0.8033$\pm$0.0216} \\
            & N & 0.6098$\pm$0.0012 & 0.5712$\pm$0.0010 & 0.6092$\pm$0.0039 & 0.6270$\pm$0.0166 & 0.6108$\pm$0.0051 & \textbf{0.6458$\pm$0.0005} & \textbf{0.6829$\pm$0.0313} \\ \hline
            
            \multirow{2}{*}{\textbf{STL-10}}  & A & 0.8496$\pm$0.0064 & 0.8786$\pm$0.0021 & 0.8320$\pm$0.0012 & 0.8784$\pm$0.0132 & 0.8814$\pm$0.0027 & \textbf{0.9054$\pm$0.0212} & \textbf{0.9231$\pm$0.0144} \\
            & N & 0.7747$\pm$0.0024 & 0.7748$\pm$0.0030 & 0.7261$\pm$0.0442 & 0.7794$\pm$0.0039 & 0.7858$\pm$0.0009 & \textbf{0.8095$\pm$0.0326} & \textbf{0.8647$\pm$0.0147} \\ \hline
            
            \multirow{2}{*}{\textbf{M-Image}} & A & 0.7333$\pm$0.0098 & 0.4375$\pm$0.0098 & 0.3986$\pm$0.0284& 0.3692$\pm$0.0024 & 0.4481$\pm$0.0093 & \textbf{0.8385$\pm$0.2488} & \textbf{0.8609$\pm$0.0271} \\
            & N & 0.7655$\pm$0.0126 & 0.6521$\pm$0.0062 & 0.5120$\pm$0.0080 & 0.5080$\pm$0.0071 & 0.7063$\pm$0.0009 & \textbf{0.8095$\pm$0.0369} & \textbf{0.8496$\pm$0.0168} \\
            \hline
        \end{tabular}}
        
        \label{tab:Comp_All}
    \end{table*}
    
    Here, we show the detailed comparison experiment results in Table \ref{tab:Comp_All}. Our methods demonstrated superior performance across a diverse range of datasets. On synthetic datasets like Moon and Ring, both SPP-FGC and SPP-FGC+ achieved exceptional accuracy and NMI scores, highlighting their robustness in handling complex and high-dimensional data structures. In tabular real-world datasets such as Bank, COIL20, USPS, and Letters, our approaches consistently outperformed all baseline federated clustering methods, underscoring their effectiveness in structured data environments with inherent privacy constraints. For complex image datasets, SPP-FGC+ showed substantial improvements on MNIST and Fashion-MNIST, though it exhibited lower performance on CIFAR10 and STL-10, suggesting areas for further optimization. Despite this, SPP-FGC maintained strong performance on these challenging image datasets. Additionally, both SPP-FGC and SPP-FGC+ excelled on the large-scale Mini-Imagenet dataset, achieving outstanding clustering results that significantly surpassed all baseline methods.
    
    The key factor contributing to this success is the iterative optimization process integrated into the adaptive feature embedding framework within SPP-FGC+. Unlike one-step algorithms, which may struggle with capturing the complex relationships in high-dimensional data, the iterative approach allows for progressive refinement of the feature space. This continuous adjustment enables the model to better align the distributed data representations, leading to more accurate and robust clustering outcomes, particularly evident in complex datasets like CIFAR-10, STL-10 and Mini-Imagenet.
    
    \begin{figure}[!htb]
        \centering
        \begin{subfigure}[b]{0.23\textwidth}
            \centering
            \includegraphics[width=\textwidth]{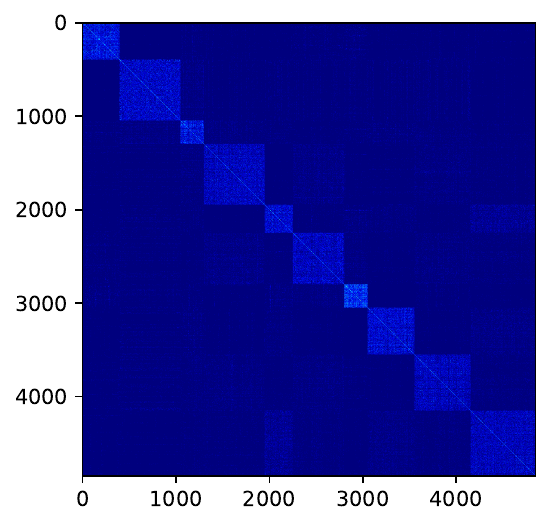}
            \caption{Private graph $E^{p,1}$}
            \label{fig:Compa}
        \end{subfigure}
        \hfill
        \begin{subfigure}[b]{0.23\textwidth}
            \centering
            \includegraphics[width=\textwidth]{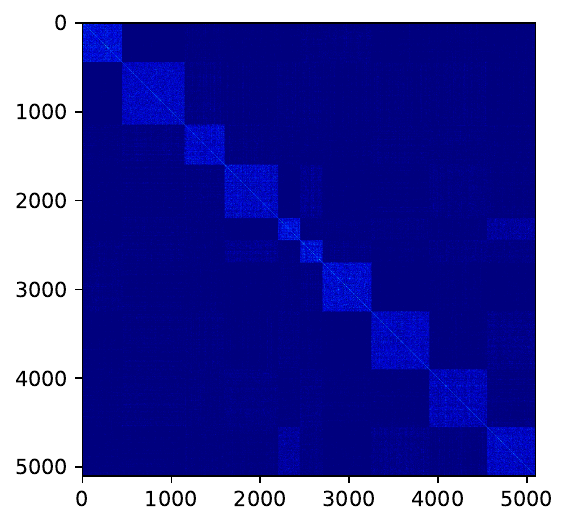}
            \caption{Private graph $E^{p,2}$}
            \label{fig:Compb}
        \end{subfigure}
        \hfill
        \begin{subfigure}[b]{0.23\textwidth}
            \centering
            \includegraphics[width=\textwidth]{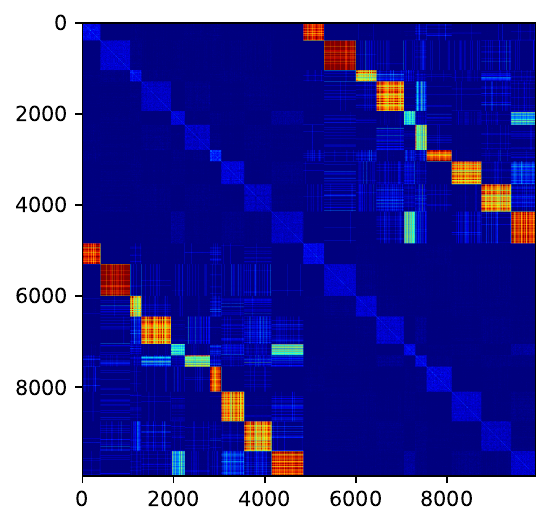}
            \caption{Global graph $E^*$}
            \label{fig:Compc}
        \end{subfigure}
        \hfill
        \begin{subfigure}[b]{0.23\textwidth}
            \centering
            \includegraphics[width=\textwidth]{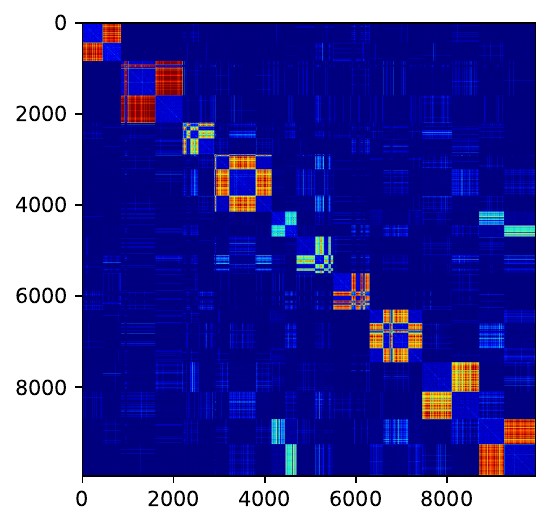}
            \caption{Aggregated graph $S$}
            \label{fig:Compd}
        \end{subfigure}
        
        \caption{Visualization of the learned structural matrices. (a) and (b) show the private graphs from two clients. (c) shows the aggregated global graph. (d) shows the final decision similarity graph.}
        \label{fig:Comp}
    \end{figure}
    
    To visualize the results of our experiment, we present Fig 5. (a) and (b) display the private structural graph learned by the SPP-FGC method on the MNIST dataset, illustrating how the local models capture the clustering structure within each client. This information is then integrated into the global graph $E^*$, shown in (c). The final global similarity graph, presented in (d), reveals a clear block diagonal structure that reflects both intra-class and inter-class relationships, highlighting SPP-FGC's effectiveness in preserving clustering integrity across the federated system.

    \subsection{F.II Heterogeneous Experiment Analysis}
    \label{APP:Het}
    
    \begin{figure}[!htb]
    \centering
    
    \begin{subfigure}[b]{0.23\textwidth}
        \centering
        \includegraphics[width=\textwidth]{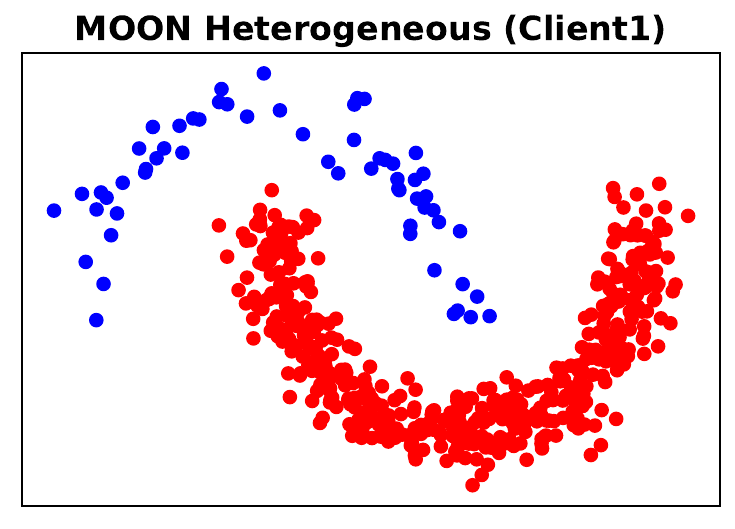}
        \caption{Client 1 data}
        \label{fig:het_client1_data}
    \end{subfigure}
    \hfill
    \begin{subfigure}[b]{0.23\textwidth}
        \centering
        \includegraphics[width=\textwidth]{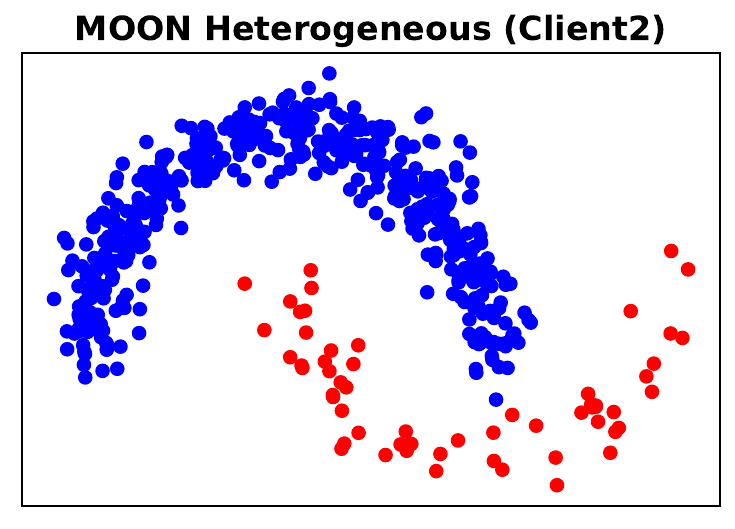}
        \caption{Client 2 data}
        \label{fig:het_client2_data}
    \end{subfigure}
    \hfill
    \begin{subfigure}[b]{0.23\textwidth}
        \centering
        \includegraphics[width=\textwidth]{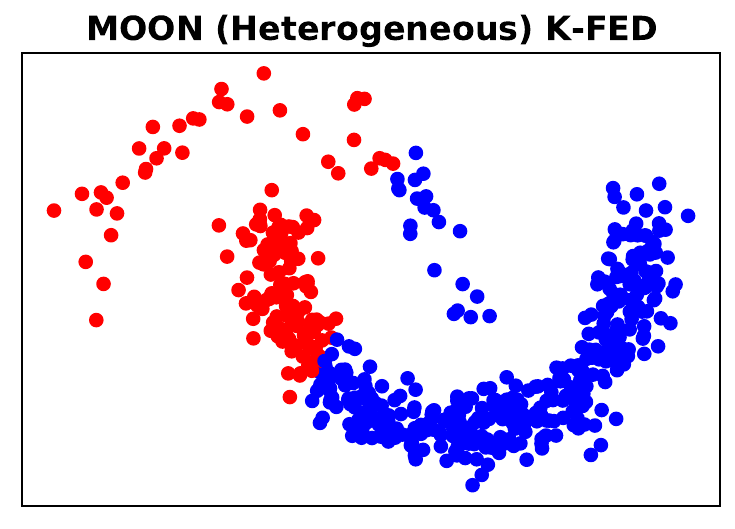}
        \caption{K-FED result}
        \label{fig:het_kfed_result}
    \end{subfigure}
    \hfill
    \begin{subfigure}[b]{0.23\textwidth}
        \centering
        \includegraphics[width=\textwidth]{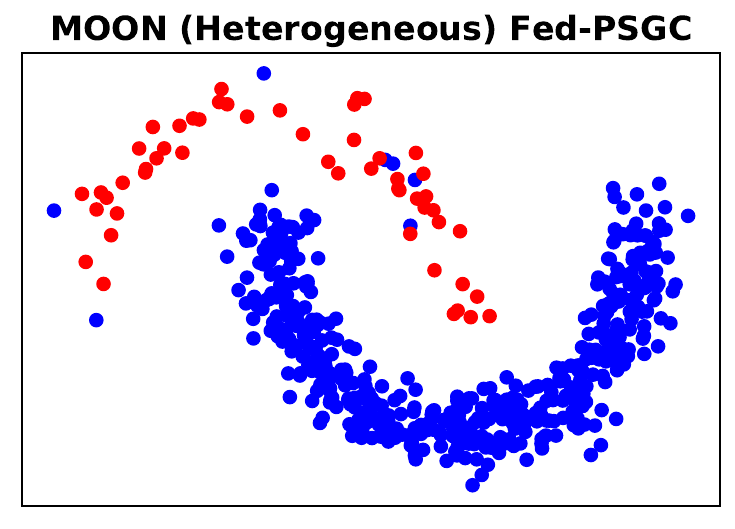}
        \caption{Our method's result}
        \label{fig:het_our_result}
    \end{subfigure}
    
    \caption{Federated clustering on heterogeneous data. (a), (b) The non-IID data distribution of the Moon dataset across two clients. (c), (d) The clustering results on Client 1's data after aggregation using K-FED and our proposed method, respectively.}
    \label{fig:Het}
\end{figure}
    
    Fig \ref{fig:Het} visualizes the federated clustering results on heterogenous data, showing the federated clustering results of K-Fed algorithm and our proposed SPP-FGC. These results indicate that our methods effectively capture the underlying clustering structure despite increasing class imbalance, maintaining clear and distinct clusters where K-FED struggles.
    
    Consider both the visualizations in Fig \ref{fig:Het} and the quantitative data in Table 2, demonstrate that our Fed-PSGC method maintains superior clustering performance compared to Fed-Kmeans and Fed-SC as data heterogeneity increases. Specifically, our method exhibits greater resilience to class imbalance, ensuring more reliable and accurate clustering even under highly heterogeneous conditions. This robustness is critical for real-world federated learning applications where data distribution across clients is often non-uniform and unpredictable.
    
    \subsection{F.III  Convergence Analysis}
    \label{APP:Conv}
    In the first experiment, we examined how the clustering performance of Fed-PSGCF scales with an increasing number of clients. Specifically, we varied the number of clients from 2 to 50, selecting client counts of 2, 3, 5, 10, 20, 30, 40, and 50. As shown in Fig 3a, Fed-PSGCF demonstrated significant improvement as the number of clients increased from 2 to 10. This initial growth phase highlights that adding more clients enhances the diversity and richness of the aggregated data, which in turn boosts clustering accuracy and robustness. However, beyond 10 clients, the performance gains plateaued, with stable clustering results maintained from 10 to 50 clients. This suggests that Fed-PSGCF effectively leverages the federated architecture to achieve consistent performance, even as the system scales, underlining its scalability and reliability across diverse federated environments.
    
    The second experiment investigated the convergence of Fed-PSGCF with respect to the number of training iterations. As depicted in Fig 3b, clustering performance steadily improved with each iteration, eventually reaching a point of convergence where additional iterations yielded minimal gains. This gradual improvement highlights the effectiveness of iterative feature learning in refining embeddings and enhancing cluster assignments over successive cycles. Fed-PSGCF’s ability to continuously enhance performance through iterative updates emphasizes its strength in deep feature extraction and refinement, critical for achieving high-quality clustering results in complex, high-dimensional data scenarios.

\end{document}